\documentclass[10pt]{article}
\usepackage{mhchem}
\usepackage{fancyhdr}
\usepackage{extramarks} 
\usepackage{amsmath}
\usepackage{amsthm}
\usepackage{amsfonts}
\usepackage{siunitx}
\usepackage{tikz}
\usepackage[plain]{algorithm}
\usepackage{algpseudocode}
\usepackage{multirow}
\usepackage{booktabs}
\usepackage{palatino}
\usepackage{graphicx}
\usepackage{subfigure}
\usepackage[colorlinks,linkcolor=black,anchorcolor=black,citecolor=black,urlcolor=blue]{hyperref}
\usepackage{amsmath,bm}
\usepackage{booktabs}
\usepackage{mathtools}
\usepackage{amssymb}
\usepackage{caption}
\usepackage{capt-of}
\usepackage{mciteplus}
\usepackage{cite}
\usepackage{mathrsfs}
\usepackage{makecell}
\usepackage[title,titletoc,toc]{appendix}
\usepackage{xr}
\usepackage{parskip}
\usepackage{soul}
\usepackage{textcomp}
\usepackage[colaction]{multicol}
\usepackage[switch]{lineno}
\usepackage{lipsum}
\usepackage{etoolbox}
\usepackage{longtable}
\usepackage{array}
\usepackage{tablefootnote}
\usepackage{graphicx}
\usepackage{color, colortbl}
\usepackage{hhline}
\usepackage[T1]{fontenc} 

\definecolor{Lightblue}{rgb}{0.867,0.914,0.961}
\definecolor{Lightgreen}{rgb}{0.883,0.934,0.848}

\newcolumntype{C}[1]{>{\centering\arraybackslash}p{#1}}
\usetikzlibrary{automata,positioning}
\usetikzlibrary{automata,positioning}

\newcommand{\crpartial}{\textup{\rmfamily\dh}}

\makeatletter
\newcommand*{\addFileDependency}[1]{% argument=file name and extension
	\typeout{(#1)}
	\@addtofilelist{#1}
	\IfFileExists{#1}{}{\typeout{No file #1.}}
}
\makeatother

\begin{document}
	
	\title{Multi-objective Molecular Optimization for Opioid Use Disorder Treatment Using  Generative Network Complex
	}
	
	\author{ Hongsong Feng$^1$, Rui Wang$^{1}$,  Chang-Guo Zhan$^{2}$, and Guo-Wei Wei$^{1,3,4}$\footnote{
			Corresponding author.		Email: weig@msu.edu} \\% Author name
		$^1$ Department of Mathematics, \\
		Michigan State University, MI 48824, USA.\\
		$^2$Deptartment of Pharmaceutical Sciences,\\
		University of Kentucky, KY 40506, USA. \\
		$^3$Department of Electrical and Computer Engineering,\\
		Michigan State University, MI 48824, USA. \\
		$^4$ Department of Biochemistry and Molecular Biology,\\
		Michigan State University, MI 48824, USA. 
	}

	\date{\today} 
	
	\maketitle

	Opioid Use Disorder (OUD) has emerged as a significant global public health issue, with complex multifaceted conditions. Due to the lack of effective treatment options for various conditions, there is a pressing need for the discovery of new medications. In this study, we propose a deep generative model that combines a stochastic differential equation (SDE)-based diffusion modeling with the latent space of a pretrained autoencoder model. The molecular generator enables efficient generation of molecules that are effective on multiple targets, specifically the mu, kappa, and delta opioid receptors. Furthermore, we assess the ADMET (absorption, distribution, metabolism, excretion, and toxicity) properties of the generated molecules to identify drug-like compounds. To enhance the pharmacokinetic properties of some lead compounds, we employ a molecular optimization approach. We obtain a diverse set of drug-like molecules. We construct binding affinity predictors by integrating molecular fingerprints derived from autoencoder embeddings, transformer embeddings, and topological Laplacians with advanced machine learning algorithms. Further experimental studies are needed to evaluate the pharmacological effects of these drug-like compounds for OUD treatment. Our machine learning platform serves as a valuable tool in designing and optimizing effective molecules for addressing OUD.
	
	
	\pagenumbering{roman}
	\begin{verbatim}
	\end{verbatim}

	\newpage
	\clearpage
	\pagebreak
	{\setcounter{tocdepth}{4} \tableofcontents}
	\newpage
	
	\setcounter{page}{1}
	\renewcommand{\thepage}{{\arabic{page}}}

	% \begin{multicols}{2}
		% \multicollinenumbers
		%\linenumbers
		\section{Introduction}

		% Brief introduction about opioid crisis or epedemic
	Opioid use disorder (OUD) is a chronic and intricate condition characterized by the compulsive seeking and use of drugs despite the detrimental effects \cite{mclellan2000drug}. It represents a significant public health concern, causing severe consequences for individuals, families, and communities. The opioid epidemic has become a pressing global health crisis, highlighting the urgent need for effective treatments for  OUD. Safe and effective medication treatments can alleviate withdrawal symptoms, reduce cravings, and help individuals maintain abstinence from opioids \cite{douaihy2013medications}.

	The main treatment methods for  OUD  typically involve a combination of medications and behavioral interventions \cite{sofuoglu2019pharmacological}, aiming to address the physical and psychological aspects of addiction, promote recovery, and prevent relapse. The U.S. Food and Drug Administration (FDA) has approved three medications including methadone, buprenorphine, and naltrexone for the treatment of OUD \cite{wang2019historical}. These medications exert their effects by binding to opioid receptors in the brain, namely mu opioid receptor (MOR), kappa opioid receptor (KOR), and delta opioid receptor (DOR). Methadone is a long-acting opioid agonist that primarily acts on MORs. It helps alleviate withdrawal symptoms and cravings \cite{fareed2010effect}. Buprenorphine, on the other hand, acts as a partial opioid agonist primarily targeting MORs. It eases withdrawal symptoms and cravings while producing less euphoria and carrying a reduced risk of respiratory depression compared to methadone \cite{bickel1995buprenorphine}. Naltrexone, classified as an opioid antagonist, blocks the effects of opioids and reduces the rewarding effects. Its mechanism of action primarily involves MORs, but it also exhibits some affinity for KORs \cite{morgan2018injectable}.

	While current medications effectively address  OUD, relapse and remission remain common due to neurobiological changes and opioid receptor tolerance resulting from repeated opioid abuse \cite{wang2019historical}. Additionally, some patients may not tolerate or respond optimally to the standard medications used for OUD. Alternative medications provide additional options to customize treatment according to individual needs and preferences. The drug discovery process encompasses several stages, including target discovery, lead discovery, lead optimization, preclinical development, and three phases of clinical trials before a new drug can be brought to market \cite{hughes2011principles}. Traditional drug discovery is a time-consuming endeavor that can extend over many years, require significant financial investments amounting to billions of dollars, and entail a substantial failure rate.

	Various methods and technologies have emerged to accelerate the drug discovery process. The number of potential drug-like molecules is estimated to be between $10^{23}$ and $10^{60}$ \cite{polishchuk2013estimation}. High throughput screening (HTS) allows for the rapid screening of large compound libraries against specific biological targets or disease models, quickly identifying leads for further medicinal chemistry optimization \cite{szymanski2011adaptation}. It allows for effective automated operation, but is associated high costs of equipment and assay development. Virtual screening involves the use of computational methods to virtually screen large databases of compounds against specific target structures. It employs molecular docking, molecular dynamics simulations, machine learning algorithms \cite{salimi2022use}. These methods enable the prediction of compound-target interactions, assessment of physicochemical and pharmacological properties, and identification of compounds with potential therapeutic effects \cite{mensa2023quantum}.

	De novo drug design (molecular generation) explores the chemical space to generate novel molecules with desirable properties. The advancement of deep learning has opened up new opportunities for innovative drug design and discovery. In recent years, many deep learning-based algorithms have been developed for de novo drug design. These algorithms utilize the power of deep learning to generate novel molecular structures with desired properties. They are trained on large datasets of known molecules, allowing them to learn the intricate patterns and relationships between chemical structures and their corresponding activities \cite{gomez2018automatic}. The machine learning algorithms used to construct these generative models can be categorized into four main types: recurrent neural network (RNN), encoder-decoder, reinforcement learning (RL), and generative adversarial network (GAN) \cite{wang2022deep}. For example, an RNN-based generative model has been proposed to generate novel molecules. By employing a fine-tuning strategy with small sets of molecules, the generated compounds can exhibit activity towards specific biological targets \cite{segler2018generating,grow2019generative} . Additionally, a variational autoencoder (VAE) has been utilized to encode molecules into a continuous latent space \cite{gomez2018automatic}. This enables operations such as generating new molecules or applying optimization strategies to design compounds with desired properties. Perturbations or interpolations can be performed on molecules' latent space vectors, and gradient-based optimization can be applied to their continuous representations \cite{gomez2018automatic}. Furthermore, a generative GAN combined with an autoencoder has been employed for de novo molecular design \cite{prykhodko2019novo}. This approach allows for the generation of random drug-like compounds or compounds biased towards specific targets. 

	In recent years, diffusion models have gained popularity in various fields, including image synthesis, video generation, and molecule design, due to their ability to produce high-quality and realistic samples. There are a few main sub-types of diffusion models: 
 latent space random noise (LSRN) \cite{grow2019generative}, 
	denoising diffusion probabilistic models (DDPMs) \cite{vignac2022digress}, score-based generative models (SGMs) \cite{niu2020permutation}, and stochastic differential equations (SDEs) \cite{huang2022graphgdp}. SDEs describe the evolution of a system over time, considering both deterministic and stochastic forces influenced by random noise \cite{zhang2023survey}. Generative diffusion models have been applied in the field of drug discovery. Random noise was introduced to latent space molecular vectors to generate novel drug-like molecules \cite{grow2019generative}.  LSRN was compared with two other approaches, latent space  controlled output  and latent space  optimized output for drug generation in a generative network complex (GNC) \cite{grow2019generative}. 
		Another example is DiffLinker \cite{igashov2022equivariant}, which utilizes diffusion models for the design of molecular linkers.  DiffSBDD \cite{schneuing2022structure}  is used in structure-based drug design to generate high-quality ligands for specific protein targets. Some recent works \cite{vahdat2021score} have also employed diffusion models trained in the latent space of autoencoders.

	In this study, we employ deep generative models to design and optimize molecules that have potential applications in the treatment of OUD. Specifically, we utilize diffusion method in the latent space of a pretrained AE model to generate novel molecules. Our objective is to create molecules that exhibit similar structural and pharmacological properties to known opioids or alternative compounds with therapeutic potential. The development of medications for OUD rely on the binding effects on opioid receptors, particularly MOR, KOR, and DOR. To achieve the molecular design, we combine a Stochastic Differential Equation (SDE)-based diffusion method with the latent space of the pretrained autoencoder model. This enables us to design molecules that are active on MOR, KOR, and DOR. In the diffusion modeling, we incorporate appropriate reference and seed compounds to steer the generation of target-biased molecules. Additionally, we employ accurate binding affinity predictors to identify potentially effective molecules that interact with these critical targets. Importantly, we consider the absorption, distribution, metabolism, excretion, and toxicity (ADMET) properties in selecting drug-like compounds for OUD treatment. By integrating ADMET criteria, we identify the generated compounds that posse desirable drug-like properties. We conduct extensive experiments to efficiently generate drug-like compounds. Furthermore, we employ a molecular optimization approach to discover additional drug candidates with nearly optimal properties. By employing different reference compounds and employing various molecular novelty thresholds, we successfully identify several drug-like compounds that exhibit multi-target effectiveness on the critical opioid receptors. Our molecular generation platform serves as a valuable tool for advancing OUD treatment.

			\section{Materials and Methods}
			
			\subsection{The structure of multi-target stochastic generative network complex (MTSGNC)}

				\begin{figure}[ht]
				\centering
				\includegraphics[width=0.65\linewidth]{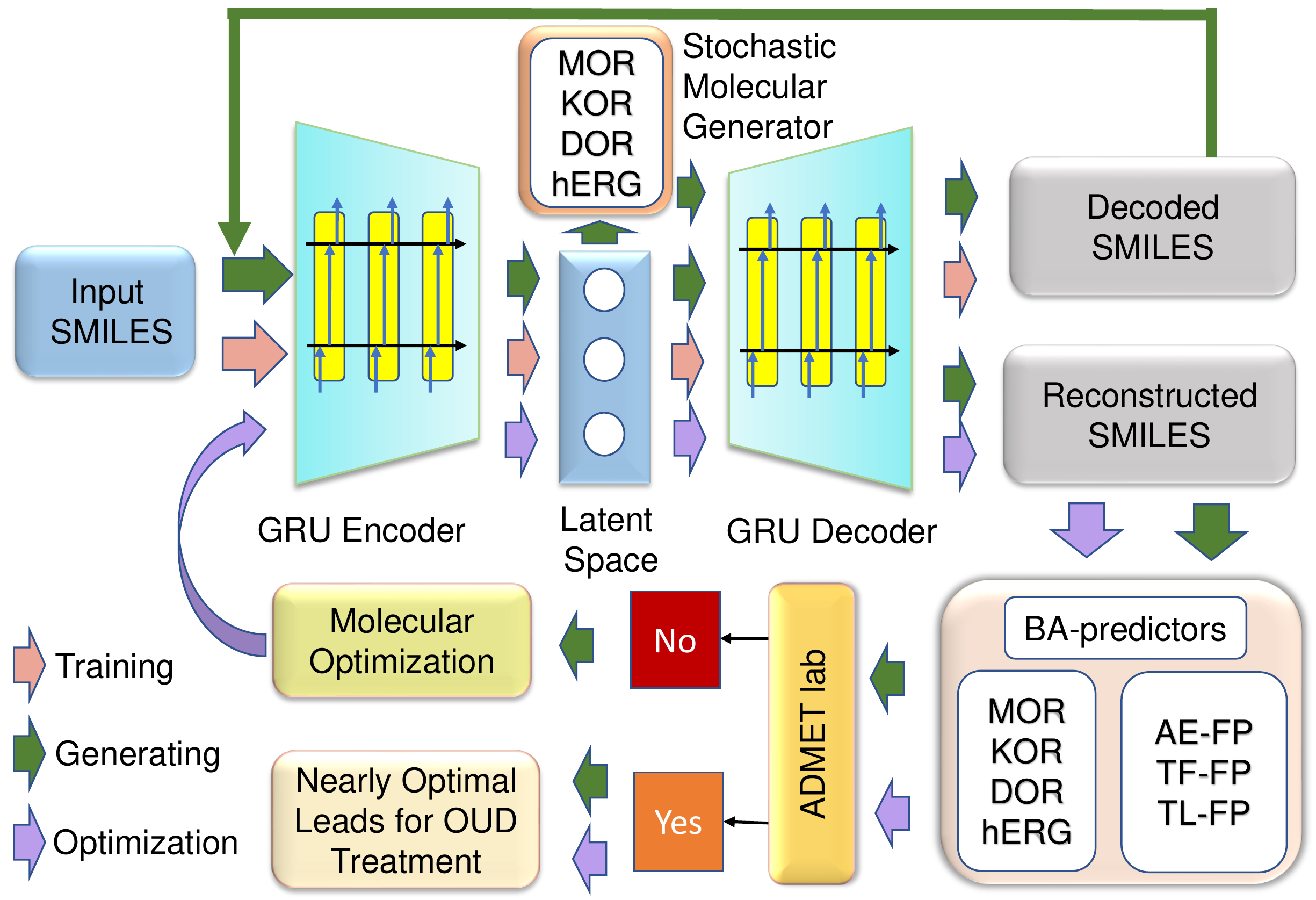} 
				\caption{{\footnotesize Schematic illustration of our multi-target stochastic generative network complex, used to design novel compounds for the treatment of  OUD. Three different paths, i.e., model training, molecular generation, and lead optimization, are colored in pink, green, and purple, respectively.} }
				\label{Fig:workflow}
			\end{figure} 
			
			\subsubsection{Sequence-to-sequence autoencoder}
			
			An autoencoder is a type of artificial neural network in deep learning that learns to compress and then reconstruct input data, such as images or text. The encoder network maps the input data to latent space, while the decoder network maps the latent space back to the original data. The compressed representation, or latent space between the two networks can capture essential features of the input data. The commonly used network for encoder and decoder can be gated recurrent unit (GRU) or a long short-term memory (LSTM) network.
			
			 In molecular science, autoencoder can be used for molecular representation learning, which involves encoding molecules into a lower-dimensional space while preserving their structural and functional properties. For instance, a sequence-to-sequence (seq2seq) autoencoder achieves this by translating one molecular string representation to another \cite{winter2019learning}. Simplified molecular-input line-entry system (SMILES) is one of the commonly used molecular representations. The dimension of the latent space is  512.  The seq2seq AE was trained with a high reconstruction ratio between inputs and output SMILES, through which latent vectors carry faithful molecular information \cite{winter2019efficient}. Therefore, the latent space can be utilized to represent the chemical space and serve as a molecular fingerprint for machine learning modeling.
			 
			 The autoencoder also has an application for generative modeling, where the decoder can be used to generate new data points by sampling from the latent space. The autoencoder (AE) structure is illustrated in the left upper part of Figure \ref{Fig:workflow}. In the current study, we utilize the pretrained seq2seq autoencoder \cite{winter2019learning} to design novel molecules for OUD treatment. 
			
			\subsubsection{Data preparation}
			
			MOR, KOR, and DOR are three critical pharmaceutical targets for treating OUD. Additionally, hERG is a crucial target that needs to be avoided in drug discovery \cite{danker2014early}. We gathered inhibitor datasets for the four targets from the ChEMBL database. The data points include SMILES strings of molecular compounds and corresponding binding labels in the form of  IC$_{50}$ or $ K_{\rm i}$. These experimental labels can be converted to binding energy with the BA=1.3633${\rm \times \log_{10}} K_{\rm i}$ (kcal/mol). As suggested by Kalliokoski \cite{kalliokoski2013comparability}, the IC$_{50}$ label can be approximated to $K_{\rm i}$ value using the relation $K_{\rm i}$=IC$_{50}/2$. The details about the datasets can be found in the Supporting Information. 
	
			The datasets are utilized in two main aspects. Firstly, the molecular compounds, particularly those exhibiting potency on MOR, KOR, and DOR, are employed as references and seeds for generating new compounds. Secondly, the four datasets are utilized as training sets to construct machine learning models for predicting binding affinity.

                \subsubsection{Stochastic molecular generator}

                Generative models serve as powerful tools for generating potential new drug molecules. In our previous work, we introduced the  GNC  for the generation of new drug-like molecules in 2019 \cite{gao2020generative,grow2019generative}. 
				We considered three latent space perturbation models, including 1) a random  noise diffusive model called ``randomized output'', 2) a gradient descending model called ``controlled output'', and 3) a multiobjective optimization model called  ``optimized output''. The first model perturbs the latent space molecular vectors with the Gaussian white noise 
and then, selects molecules that have improved properties in a manner similar to the Monte Carlo method. This model ensures the novelty of generated molecules but may not be effective in reaching the desirable drug-like properties. The second model improves a specific molecular properties of generated molecules by a force driven term. However, the resulting new molecules may not retain other important properties and lack novelty. The third approach were designed to simultaneously optimize multiple molecular properties of generated molecules via a multi-objective loss function. However, the novelty of generated molecules may not be guaranteed.  To improve the performance of our GNC model,  we propose to combine our 	random  noise diffusive model with 	our multi-objective optimization model. The resulting model can be regarded as a drift and diffusion model as described by the Langevin equation.    						
%								decide to incorporate the idea of diffusion-based models \cite{xu2022geodiff, corso2022diffdock}, which have gained substantial popularity following the release of platforms such as DALL-E by OpenAI in 2021.

                Langevin equation is a stochastic differential equation (SDE), which is used to describe the diffusion processes such as the the random motion of the particles over time in the particle's velocity space, taking into account of both deterministic forces and random forces. It can be regarded as stochastic generalization of the Newton's equation of motion.  In this work, one of our goals is to employ Langevin equation to optimize the molecular generator in our previous GNC model.  
                
                We assume $\mathbf{X}$ is a latent space vector of a molecule with 512 dimensions, and $\mathbf{X}_k$ represents its $k$th latent space reference vector. Then the Langevin equation of our drug generator system is:
                \begin{equation} \label{eqn:Langevin}
                    \frac{d\mathbf{X}}{dt} = \alpha \sum_k a_k(\mathbf{X}_k - \mathbf{X}) + \bm{\xi}(t),
                \end{equation}
                where $a_k$ is a positive weighting parameter corresponds to $\mathbf{X}_k$ satisfying $\displaystyle \sum_k a_k =1$, $\bm{\xi}(t)$ is a Gaussian white noise, and $\alpha$ is a hyperparameter. The first term in Eq. (\ref{eqn:Langevin}) is a force term  which gives rise to a gradient descent towards   targets $\mathbf{X}_k$.  
  		Then according to  Eq. (\eqref{eq:1d Langevin eq sol}) in Section \ref{subsubsec:langevin}, the general solution of this system is given by:
                \begin{equation}\label{eq:1d Langevin eq}
                    \mathbf{X}(t) = \mathbf{C}^{-\alpha t} + \int_0^t e^{-\alpha(t-u)} (\alpha\sum_k a_k\mathbf{X}_k + \bm{\xi}(u))du,
                \end{equation}
                where the initial state $\mathbf{X}(0) = \mathbf{C}$. The latent vector $\mathbf{C}$ is for an initial seed molecule.

                It is worth noting that the Langevin equation provides a molecule-wise description of   drift and diffusion processes. However, to obtain an understanding of how the overall distribution of molecules evolves over time, a statistical perspective is needed. Therefore, we also discuss the Fokker-Planck equation derived from the Langevin equation (the detailed derivation can be found in Section \ref{subsubsec:FokkerPlanck}), providing a connection between the dynamics of individual drug-like molecules and the statistical behavior of the entire system. The Fokker-Planck equation facilitates the GAN-based generation of drug molecules. 
                
			\subsubsection{Multi-target stochastic molecular generation}			
			
			Figure \ref{Fig:workflow} presents the compound generation process in our  GNC. We aim at designing novel drug-like compounds effective on MOR, KOR, and DOR, while having no hERG side effect. The molecular generation process comprises four key steps, which are further elaborated in the subsequent subsections.
						
			\begin{enumerate}
				\item  Select three compounds that are potent on MOR, KOR, and DOR respectively, as reference compounds from the collected datasets. Also Pick a seed compound that is potent on multiple of the three receptors. Then encode the SMILES strings of the references and seed compounds into latent vectors through the pretrained encoder.  
				
				\item Input the latent vectors of reference and seed compounds  into the stochastic molecular generator, giving rise to a large number of new latent vectors. These new latent vectors, as the representations of potential molecules, are screened with a constraint of binding affinity $\Delta G <$ -9.54 kcal/mol $(K_{\rm i}=1 \rm nm)$ on MOR, KOR, and DOR, as well as $\Delta G >$ - 8.18 kcal/mol $(K_{\rm i}=10 \rm nm)$ on hERG. Pretrained BA predictors are used in the BA constraint evaluation.

				\item Latent vectors with desired BA properties are decoded into valid SMILES (interpretable by RDKit). These SMILES are subsequently fed back through the pretrained encoder and decoder again, as illustrated in Figure \ref{Fig:workflow}, to identify molecules that can be successfully reconstructed. Those reconstructed molecules are considered to be structurally stable and well-interpreted by the autoencoder networks.
				
				\item Those reconstructed molecules are reevaluated on their binding affinities and ADMET properties to identify drug-like compounds. A molecular optimization process can be implemented to design more drug candidates with desired druggable profiles. 
				
			\end{enumerate}

			To generate novel molecules, different references and seed compounds, or weights for reference latent vectors can be used in Step 2. The hyperparameters $\alpha$ in Eq. (\ref{eqn:Langevin}) is set as 0.15 in this study. As we consider three reference compounds corresponding to the three receptors, we apply three reference vectors in Eq. (\ref{eqn:Langevin}). Weight coefficients $\alpha_k$, $k=1,2,3$, are for MOR, KOR, and DOR reference compounds, respectively. Different weight values can be used to emphasize the importance of selected references. The parameter $t$ in Eq. (\ref{eq:1d Langevin eq}) can be various numbers to generate vectors for potential molecules.

			\subsection{Binding affinity predictors}			
			
			Within the generator of our GNC, we incorporate four BA predictors to assess the BA of the generated potential molecules on the four crucial targets. To construct these predictors, we utilize molecular fingerprints derived from the latent space of an autoencoder. These fingerprints are referred to as AE-FP, and the resulting four BA predictors are denoted as AE-BPs. The AE-BPs were fulfilled by integrating the AE-FPs with a deep neural network (DNN) algorithm.

			Consensus models are used to further evaluate the BAs of those reconstructed SMILES, as shown in the right bottom of Figure \ref{Fig:workflow}. Two more molecular fingerprints, namely, transformer fingerprint (TF-FP) and topology Laplacian (TL-FP) fingerprint are used. They were designed with a pretrained transformer model \cite{chen2021extracting} and our recently proposed topology Laplacian theory\cite{wang2020persistent}. Brief descriptions about the two fingerprints are provided in the following subsections.
			 
			 BA-predictors were constructed by integrating the TF-FP with a deep neural network (DNN) algorithm, and fusing the TL-FP with a gradient boosting decision tree (GBDT) algorithm. We refer to these two models as TF-BP and TL-BP, respectively. The consensus model or prediction is obtained by averaging the predictions from AE-BP, TF-BP and TL-BP. This strategy significantly enhances machine learning predictions \cite{feng2023virtual,gao20202d} and typically outperforms individual models.
			 
			Both DNN and GBDT algorithm are popular algorithms in building machine learning models. DNN have advantages of dealing large and complex datasets, constructing hierarchical features and modeling complex nonlinear relationships. GBDT as an ensemble algorithm has merits of being less sensitive to hyperparameters, less prone to overfitting, and easy to implement. In building our machine-learning BA predictors, AE-FP and TF-FP showed better predictive performance when combined with DNN, while TL fingerprints has better predictive ability when integrated with GBDT. The details about the hyperparameter for AE-BPs, TF-BPs, and TL-BPs can be found Table S2 in the Supporting Information. 
			
			The predictive performance of the consensus models was evaluated using five-fold cross-validation. The average Pearson correlation coefficients (R) obtained were 0.824, 0.840, 0.845, and 0.756 for the MOR, KOR, DOR, and hERG datasets, respectively. Additionally, the average root-mean-squared error (RMSE) values were 1.010, 1.027, 1.006, and 0.801 kcal/mol for the same datasets.

			\subsubsection{Topological Laplacian molecular fingerprint}
									
			In this subsection, we give a brief explanation of topological Laplacians (TL) based on spectral graph theory \cite{wang2020persistent}. Topology offers significant simplification of biomolecular data by dealing with the connectivity of different components in a space, and characterizes independent entities, rings and higher dimensional faces within the space \cite{mischaikow2004computational,zomorodian2004computing}. It can be used for a high level of abstraction to three-dimensional (3D) biomolecular structures. Topological Laplacian can reveal both topological invariants and homotopic shape information through the harmonic and non-harmonic spectra of the Laplacian matrix. Intricate shape information can be obtained through evolving manifolds defined under filtration parameters. The topological space is based on geometric components of a data set, including discrete vertices, edges, triangles, tetrahedrons in the context of 3D molecular structures. TL form families of persistent $q$-combinatorial Laplacian operators, providing a powerful multiscale analysis tool. These operators are derived from persistent spectral graph theory, as illustrated below.

			 The persistent Laplacians are defined under a filtration of an oriented simplicial complex $K$. A sequence of the sub-complexes $\{K_t\}_{t=0}^m$ of $K$ is constructed
			\begin{align*}
				\emptyset =K_0 \subseteq K_1 \subseteq K_2 \subseteq \cdots \subseteq K_m =K.
			\end{align*}
		
			On each simplicial complex $K_t$, a chain complex is defined as $C_q^t:=C_q(K_t)$ and there exists a $q$-boundary operator $\partial_q^t: C_q(K_t)\rightarrow C_{q-1}(K_t)$. For the general case with $0<q\leq$ dim$(K_t)$, the $q$-boundary operator is in the following form
			\begin{align*}
				\partial_q^t(\sigma_q)=\sum\limits_i^q(-1)^i\sigma_{q-1}^i,\enspace\text{for}\enspace \sigma_q\in K_t,
			\end{align*}
			where $\sigma_q=[v_0,v_1,\cdots,v_q]$ is an oriented $q$-complex, and $\sigma_{q-1}^i=[v_0,\cdots,\hat{v}_i,\cdots,v_q]$ is an oriented $(q-1)$-simplex by removing vertex $v_i$. For the case of $q<0$, the $C_q(K_t)=\{\emptyset\}$ and $\partial_q^t$ is a zero map. The $q$-adjoint boundary operator is defined as the adjoint operator that corresponds to the $q$-boundary operator.
			\begin{align*}
				\partial_q^*:C_{q-1}(K_t)\rightarrow C_{q}(K_t).
			\end{align*}
			We consider $\mathbb{C}_q^{t+p}$, a subset of $C_q^{t+p}$ with its boundary in $C_{q-1}^t$:
			\begin{align*}
				\mathbb{C}_q^{t+p}:=\{\sigma \in C_q^{t+p}| \partial_q^{t+p}(\sigma)\in C_{q-1}^t\}.
			\end{align*}
			For this subset, the $p$-persistent $q$-boundary operator  $\crpartial_q^{t+p}:\mathbb{C}_q^{t+p}\rightarrow C_{q-1}^t$ and the adjoint boundary operator $(\crpartial_q^{t+p})^*:C_{q-1}^{t}\rightarrow \mathbb{C}_{q}^{t+p}$ are well defined. The $p$-persistent $q$-combinatorial Laplacian operator is given as 
			\begin{align*}
				\Delta_q^{t+p}=\crpartial_{q+1}^{t+p}\left(\crpartial_{q+1}^{t+p}\right)^*+(\partial_q^t)^*\partial_q^t,
			\end{align*}
			together with its matrix representation as 
			\begin{align*}
				\mathcal{L}_q^{t+p}=\mathcal{B}_{q+1}^{t+p}\left(\mathcal{B}_{q+1}^{t+p}\right)^T+\left(\mathcal{B}_q^t\right)^T\mathcal{B}_q^t.
			\end{align*}
			Matrices $\mathcal{B}_{q+1}^{t+p}$ and $\mathcal{B}_{q}^{t}$ are the matrix representations for boundary operators $\crpartial_{q+1}^{t+p}$ and $\crpartial_{q}^{t}$, respectively. The row number of $\mathcal{B}_{q+1}^{t+p}$ is equal to the number of oriented $q$-simplices in $K_t$, and the column number equal that of oriented $(q+1)$-simplices in $K_{t+q}\cap \mathbb{C}_{q+1}^{t+p}$. In addition, the transposes of $\mathcal{B}_{q+1}^{t+p}$ and $\mathcal{B}_{q}^{t}$ are the matrix representation for $\left(\crpartial_{q+1}^{t+p}\right)^*$ and $(\partial_q^t)^*$. The topological and spectral information of $K_t$ can be accessed from the Laplacian operator. We denote spectra of $\mathcal{L}_q^{t,p}$ as a set
			\begin{align*}
				\text{Spectra}\left(\mathcal{L}_q^{t+p}\right) = \left\{(\lambda_1)_q^{t+p}, (\lambda_2)_q^{t+p},\cdots, (\lambda_N)_q^{t+p}\right\},
			\end{align*}
			where $N$ indicates the dimension of $\mathcal{L}_q^{t+p}$. 
			The Betti numbers, the number of zero eigenvalues, of $\mathcal{L}_q^{t,p}$ can reveal $q$-cycle information. For the $p$-persistent $q$-combinatorial Laplacian matrix $\mathcal{L}_q^{t+p}$, the Betti number is defined as
			\begin{align*}
				\beta_q^{t+p}=\text{dim}\left(\mathcal{L}_q^{t+p}\right)-\text{rank}\left(\mathcal{L}_q^{t+p}\right)=\text{nullity}\left(\mathcal{L}_q^{t+p}\right)=\text{number of zero eigenvalues of }\mathcal{L}_q^{t+p}.
			\end{align*}
			The $\beta_q^{t+p}$ value indicates the number of $q$-cycles in simplices $K_t$ that are still alive in simplices $K_{t+p}$. For the biomolecular data, the order of $q$ ranges from 0 up to 2, as the data is in three dimensional space. The values of $\beta_q^{t+p}$ measures the persistence of connected components, tunnels or circles, and cavities or voids. The harmonic persistent spectra track the topological changes while non-harmonic persistent spectra records the geometric changes.
			
			Based on aforementioned topological Laplacians, we form a set of molecular features by using the eigenvalue statistics of Laplacian matrix $\mathcal{L}_0^{t+p}$. The features are compromised of $\beta_0^{t+p}$ and the sum, mean, median, maximum, minimum, standard deviation, variance, sum of the square of the non-harmonic spectra. The representability of TL feature for molecules depends on the selection of atoms with different combinations of elements, which in turn construct distinct oriented $q$-simplices in $K_t$. As a result, element-specific Laplacian matrices are defined under a set of filtration. To enhance the representability of TL features, it is necessary to analyze the given dataset on element types and atomic proportions. Additionally, this study utilizes a filtration radius with a lower bound of 1 angstrom and an upper bound of 10 angstroms. This range is chosen based on the observation that the majority of compounds in each dataset have a three-dimensional size of less than 20 angstroms in each Cartesian direction. More details of element-specific topological Laplacians based on distribution analysis can be found in the Supporting information.
			
			\subsubsection{Bidirectional transformer molecular fingerprint}
			
			In a recent work \cite{chen2021extracting}, a self-supervised learning (SSL) platform was developed to pretrain deep learning models on millions of unlabeled molecules. This platform generated latent space vectors for input SMILES \cite{chen2021extracting}. The pretraining of SSL was accomplished by employing the bidirectional encoder transformer (BET) model. Within the SSL pretraining platform, SMILES strings were encoded by constructing pairs of real SMILES and masked SMILES, with a certain percentage of symbols in the strings hidden. The model was then trained in a supervised manner using these data-mask pairs \cite{chen2021extracting}. The attention mechanism was utilized to capture the significance of each symbol in the SMILES strings. A set of molecular fingerprint can be obtained by averaging 256 embedding vectors associated with a given SMILES string. For the training of the SSL-based BET model, molecular SMILES from ChEMBL databases were employed, and the latent vector transformer fingerprints (TF-FP) generated by the pretrained model were used as molecular fingerprints in this study.

		\subsection{Langevin and Fokker-Planck equations}
		
            \subsubsection{Random variables and expected value} A random variable $X$ is a variable whose possible values are outcomes of a random phenomenon. The random variable can be either discrete, taking on a countable number of values, or continuous, taking on any value within a certain range or set. For a discrete random variable, we can write $P(X=x)$ to denote the probability that $X$ takes the value $x$. The expected value of random variable $X$ is $\displaystyle E(X) = \sum_{i}x_iP(X=x_i)$. For a continuous random variable, we talk about probability density function (pdf) $p(x)$ such that for any interval $[a, b]$, $\displaystyle P(a \leq X \leq b) = \int_{a}^{b} p(x) \, dx$. Then the expected value of random variable $X$ is $\displaystyle E(X) = \int_{-\infty}^{+\infty}xp(x)dx$.
            
            \subsubsection{Langevin equation} \label{subsubsec:langevin}
            The Langevin equation is a commonly used stochastic differential equation (SDE) in physics that aims to describe the behavior of a system as it evolves over time under the influence of deterministic drift and random (fluctuating) forces. The Langevin equation can describe the motion of a particle in a fluid,
            \begin{equation}
                m\frac{d\mathbf{v}}{dt} = -\lambda \mathbf{v} + \mathbf{\eta}(t),
            \end{equation}
            where $m$ is the mass of the particle, $\mathbf{v}$ is the velocity of the particle, $\lambda$ is its corresponding damping coefficient, and $\mathbf{\eta}$ is the noise term which represent the effect of the collisions with the molecules of the fluid. In many cases, the 1-dimensional Langevin equation is   written in a general form as:
            \begin{equation}
                \frac{dx}{dt} = - \gamma x + \xi(t),
            \end{equation}
            where $\xi(t)$ is a Gaussian white noise process with $\langle \xi(t) \rangle = 0$ and $\langle \xi(t) \xi(t') \rangle = \delta(t - t')$. The general solution of the 1-dimensional Langevin equation has the form:
            \begin{equation}\label{eq:1d Langevin eq sol}
                x(t) = Ce^{-\gamma t} + \int_0^t e^{-\gamma(t-u)}\xi(u)du,
            \end{equation}
            where the initial state $x(0) = C$.

            % \subsection{Derivation of Fokker-Planck equation from stochastic differential equation}
            \subsubsection{Connections between Fokker-Planck equation and Langevin equation}\label{subsubsec:FokkerPlanck}
            For an It\^{o} process driven by the standard Wiener process $W_t$ in one spatial dimension $x$, a SDE  is given in the form
            \begin{equation}\label{eq:sde}
                dX_t = \mu(X_t, t) dt + \sigma(X_t, t) dW_t
            \end{equation}
            where $\mu_t = \mu(X_t, t)$ and $\sigma_t = \sigma(X_t, t)$ are given functions representing the drift and diffusion coefficients, respectively. To derive the Fokker-Planck equation (also called forward kolmogorov equation), we can first apply It\^{o}' lemma. Assume $f(x)$ is a twice-differential function, then its expansion in Taylor's series is:
            \[
            df = f_x dX_t + \frac{1}{2} f_{xx} dX_t^2 + \dots.
            \]
            Substituting the Eq. \eqref{eq:sde} into this, we have:
            \begin{align}\label{eq:Ito Lemma}
                df & = f_x(\mu_t dt + \sigma_tdW_t) + \frac{1}{2} f_{xx} (\mu_t dt + \sigma_t dW_t)^2 + ...\\
                   & = f_x\mu_t dt + f_x \sigma_tdW_t + \frac{1}{2}f_{xx} (\mu_t^2 dt^2 + 2\mu_t\sigma_tdtdW_t + \sigma_t^2dW_t^2).
            \end{align}
            In the limit $dt \to 0$, the terms $dt^2$ and $dt W_t$ tend to zero faster than $dW_t^2$. Meanwhile, since $W_t$ is a Wiener process, which assures $dW_t^2 = dt$. Therefore, from Eq. \eqref{eq:Ito Lemma}, we can obtain 
            \[
            df = (\mu_tf_x + \frac{1}{2}f_xx\sigma_t^2)dt + \sigma_tf_xdW_t,
            \]
            then its expected value would be:
            \begin{align*}
                E(df) & = E[(\mu_tf_x + \frac{1}{2}f_{xx}\sigma_t^2)dt + \sigma_tf_xdW_t]\\
                      & = E[(\mu_tf_x + \frac{1}{2}f_{xx}\sigma_t^2)dt] + E[\sigma_tf_xdW_t]\\
                      & = E[(\mu_tf_x + \frac{1}{2}f_{xx}\sigma_t^2)dt] + 0.\\
            \end{align*}
            Rearranging terms, we can get:
            \[
            \frac{dE[f]}{dt} = E[\mu_tf_x + \frac{1}{2}f_{xx}\sigma_t^2].
            \]
            Let $p(x,t)$ to be the probability density function of the random variable $x$. Then by applying the definition of expected value, we have:
            \begin{align*}
                \frac{d}{dt}\int_{-\infty}^{+\infty} f(x)p(x,t)dx 
                & = \int_{-\infty}^{+\infty}\left(\mu_tf_x  + \frac{1}{2}f_{xx}\sigma_t^2\right)p(x,t) dx\\
                & = \int_{-\infty}^{+\infty}\mu_tf_xp(x,t)dx + \frac{1}{2}\int_{-\infty}^{+\infty}f_{xx}\sigma_t^2p(x,t) dx\\
                & = -\int_{-\infty}^{+\infty}f(x)\mu_t\frac{\partial p}{\partial x}dx - \frac{1}{2}\int_{-\infty}^{+\infty}\sigma_tf_x\frac{\partial p}{\partial x}dx\\
                & = -\int_{-\infty}^{+\infty}f(x)\mu_t\frac{\partial p}{\partial x}dx + \frac{1}{2}\int_{-\infty}^{+\infty}f(x)\sigma_t^2\frac{\partial^2 p}{\partial x^2}dx\\
                & = \int_{-\infty}^{+\infty}f(x) \left(-\mu_t\frac{\partial p}{\partial x} + \frac{1}{2}\sigma_t^2\frac{\partial^2 p}{\partial x^2} \right)dx.
            \end{align*}
            Therefore, one can get the Fokker-Planck equation of the probability density function $p(x,t)$ of the random variable $X_t$ in the one spatial dimension:
            \begin{equation}
                \frac{\partial p}{\partial t} = -\mu_t\frac{\partial p}{\partial x} + \frac{1}{2}\sigma_t^2\frac{\partial^2 p}{\partial x^2}
            \end{equation}
  The   Fokker-Planck equation describes the time evolution of the statistical distribution of a forced particle in a dissipative media. It can be used to generate drug-like  molecules via a GAN system.

			\section{Experiments}
			
			\subsection{Generating novel mutil-target inhibitors for MOR, KOR, and DOR}
			
		\begin{figure}[ht]
			\centering
			\includegraphics[width=0.95\linewidth]{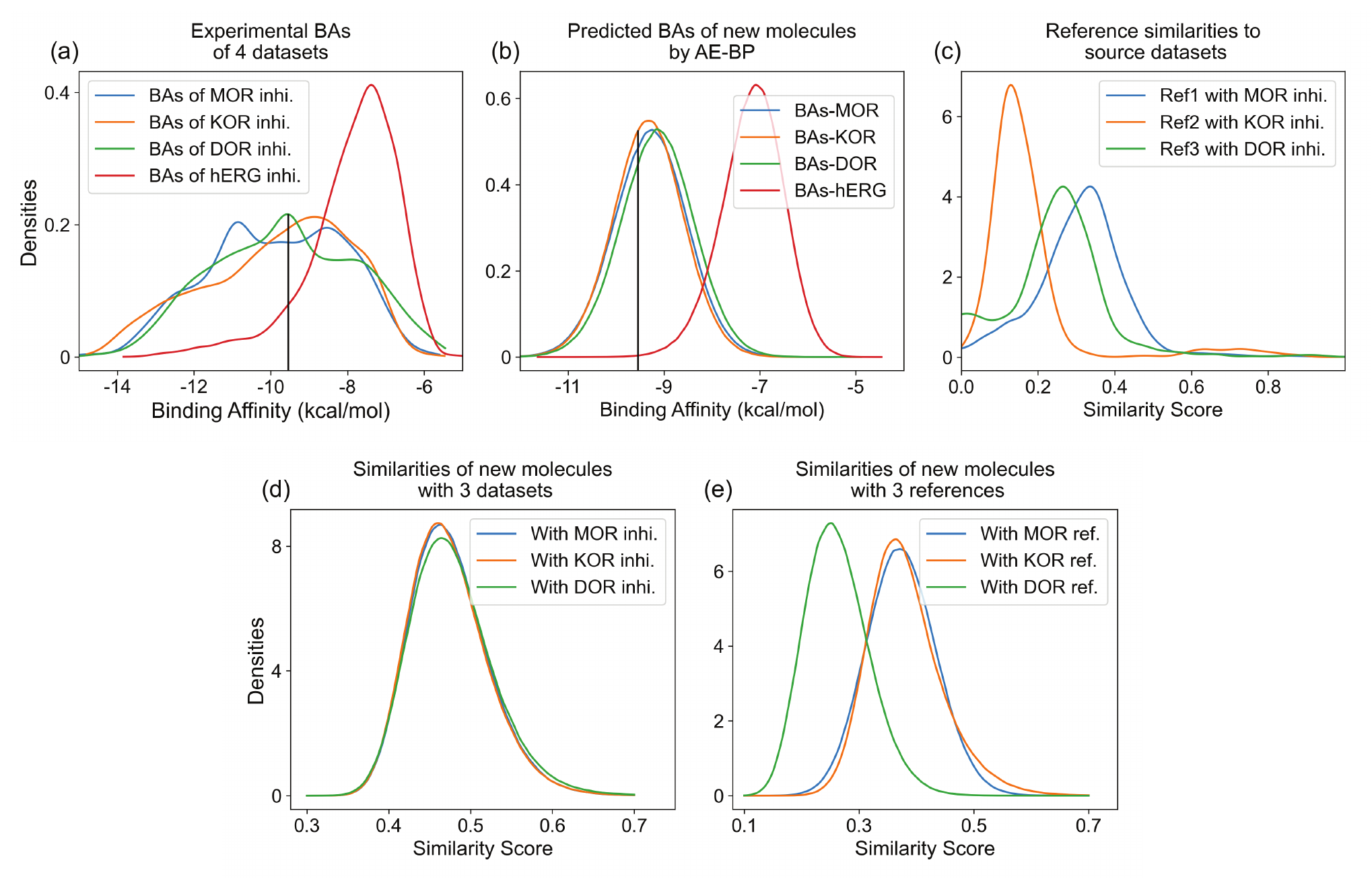} 
			\caption{{\footnotesize Analysis on binding affinity (BA) distribution of the training data and the generated compounds, as well as the analysis on similarity score distributions. a.    BA distributions of the inhibitors in the four training datasets. The unit for BA is kcal/mol. b. BA distributions of the generated molecules predicted by AE-BPs. c. The distributions of similarity scores between reference compounds and corresponding source inhibitor dataset. d. The distributions of similarity scores between generated compounds and inhibitor datasets for the three opioid receptors. e. The distributions of similarity scores between generated compounds and three reference compounds.
			} }
			\label{Fig:distributions-BA-similarity}
		\end{figure} 
	
			The FDA-approved medications for OUD treatment are highly effective on MOR, KOR, and DOR. We utilize our GNC to design more molecules that are simultaneously effective on the three receptors. Meanwhile, we avoid those inhibitors with potential hERG side effects. 
			
			Selecting appropriate reference and seed compounds are crucial in generating effective molecules. We prioritize compounds from the collected datasets that show effectiveness on the opioid receptors as our references. This is because the generated compounds, inheriting the pharmacophores of such references, are more likely to exhibit potency on the receptors. Figure \ref{Fig:distributions-BA-similarity}a displays the  binding affinity (BA)  distributions of these datasets. It is evident that there is a significant number of effective molecules in the MOR, KOR, and DOR datasets, all having binding affinity values below -9.54 kcal/mol. This threshold is widely accepted for identifying active compounds. The three datasets collectively consist of 2152 common compounds, from which we select our reference or seed compounds
			
			We rely on machine learning models to predict the BA values of the generated compounds. The three datasets demonstrate broad BA distributions, spanning from -14 to -6 kcal/mol, which indicates the presence of highly diverse molecules. Moreover, the BA data exhibit balanced distributions in relation to the BA threshold of -9.54 kcal/mol. This balanced distribution of training data enables unbiased BA prediction.
			 
			 Each dataset contains a moderate number of molecules with  BAs  ranging between -12 kcal/mol and -10 kcal/mol. We prioritize selecting compounds within this range as reference or seed compounds for two reasons. Firstly, these compounds increase the likelihood of generating potent molecules. References showing high effectiveness on multiple targets are especially valuable for this purpose. Secondly, the ample data within this BA range aids in accurately identifying potent inhibitors through machine learning predictions..

			Reference compounds are pivotal in drug design as they greatly impact the novelty of generated compounds. Two crucial factors influencing novelty are the number of reference compounds and the coefficient weights assigned to their AE latent vectors. Novelty is measured by comparing the similarities between the generated compounds and the reference compounds, with lower similarity indicating higher novelty. In the upcoming experiments, we will demonstrate the mechanism of our molecular  GNC and specifically investigate the impact of reference numbers on molecular novelty.

			\subsubsection{Generation using three distinct reference molecules}
			 
			 In our first experiment, we selected three compounds, namely ChEMBL2048770, ChEMBL3349979, and ChEMBL494462, from the inhibitor datasets. Each of these compounds demonstrates effectiveness on MOR, KOR, and DOR, with corresponding  BA  values of -11.51, -11.58, and -11.92 kcal/mol, respectively. Therefore, we utilized them as reference compounds for each respective receptor. The seed compound exhibits BA values of -10.44, -10.5, and -8.96 kcal/mol on MOR, KOR, and DOR, respectively. It acts as a weak inhibitor on DOR. By using ChEMBL494462 as the reference compound for KOR, the generated compounds can incorporate certain moieties present in ChEMBL494462. Consequently, those generated molecules may exhibit effectiveness on DOR. When applying the three references in the molecular generator, the weight coefficients $(\alpha_1, \alpha_2, \alpha_3)$ in Eq. (\ref{eqn:Langevin}) are set to $(0.35, 0.35, 0.3)$.

			 Using the reference and seed compounds, our GNC generated over a million novel and valid molecules in just a few hours using supercomputers. Subsequently, these compounds were passed through the encoding-decoding network, and we retained those that could be successfully reconstructed for further  BA  reevaluation and ADMET analysis. The reconstruction rate of the generated compounds was $90.1\%$, yielding to a vast library of novel molecules.

			  In the molecular generation process, we use AE-BPs to evaluate the BAs of the generated compounds. Those multi-target active molecules are identified. Figure \ref{Fig:distributions-BA-similarity}b indicates the BA distributions of the reconstructed molecules by our AE-BPs. A large number of the generated molecules can be effective on each of the three opioid receptors. In addition, only a very small portion of these molecule can cause hERG side effect. It is promising to find enough multi-target active compounds.

			 Investigating molecular similarity scores from various perspectives is crucial as they are associated with machine learning predictions. We are interested in examining the similarity scores between three reference compounds and their respective source datasets. These reference compounds are selected from the MOR, KOR, and DOR inhibitor datasets, which were utilized as training data for developing BA predictors. The generated molecules exhibit similarities with the reference compounds. Figure \ref{Fig:distributions-BA-similarity}c presents the distribution of similarity scores between the reference compounds and their source datasets. A small number of molecules within each training dataset exhibit high similarities with the corresponding reference compounds. The similarity scores are calculated using Tanimoto coefficients, comparing the AE latent vectors of the reference compounds with the molecules in each dataset. On average, the similarity scores between each reference compound and its corresponding dataset remain below 0.4. However, there are still more than ten molecules in each dataset that exhibit similarity scores over 0.85 for the selected reference compound.

			  Figure \ref{Fig:distributions-BA-similarity}d showcases the distribution of similarity scores between the generated molecules and the three inhibitor datasets. Each similarity score is determined by the highest Tanimoto coefficient calculated between the AE latent vectors of a generated molecule and all molecules in the respective dataset. The majority of similarity scores fall within the range of 0.4 to 0.6, indicating high levels of novelty among the generated molecules. This outcome aligns with expectations, as each new molecule incorporates molecular features from three distinct reference compounds, resulting in unique molecular structures. Notably, the distribution curves of the three datasets exhibit a consistent pattern.

			  Figure \ref{Fig:distributions-BA-similarity}e presents the distribution of similarity scores between the generated molecules and the three reference compounds. The majority of similarity scores are found to be below 0.5, further confirming the high novelties of the generated molecules. The similarity distributions of the molecules with MOR and KOR reference compounds demonstrate a consistent pattern, while the average similarity scores with the DOR reference compound are comparatively lower than those for MOR and KOR. This discrepancy is primarily attributed to the weights assigned to Eq. (\ref{eqn:Langevin}), specifically $(0.35, 0.35, 0.3)$. A higher weight assigned to a reference compound leads to a greater similarity between the generated molecules and the designated reference compound.

			 The novelties of the generated molecules are observed in Figure \ref{Fig:distributions-BA-similarity}d and \ref{Fig:distributions-BA-similarity}e. High novelties comes with the risk of inaccurate BA prediction, as a higher molecular similarity with the training data can have more reliable predictions. To address this issue, we can either use more accurate BA-predictors or reduce the molecular novelty.

			  \subsubsection{Generation using two reference molecules}
			  
			  	\begin{figure}[ht]
			  	\centering
			  	\includegraphics[width=0.72\linewidth]{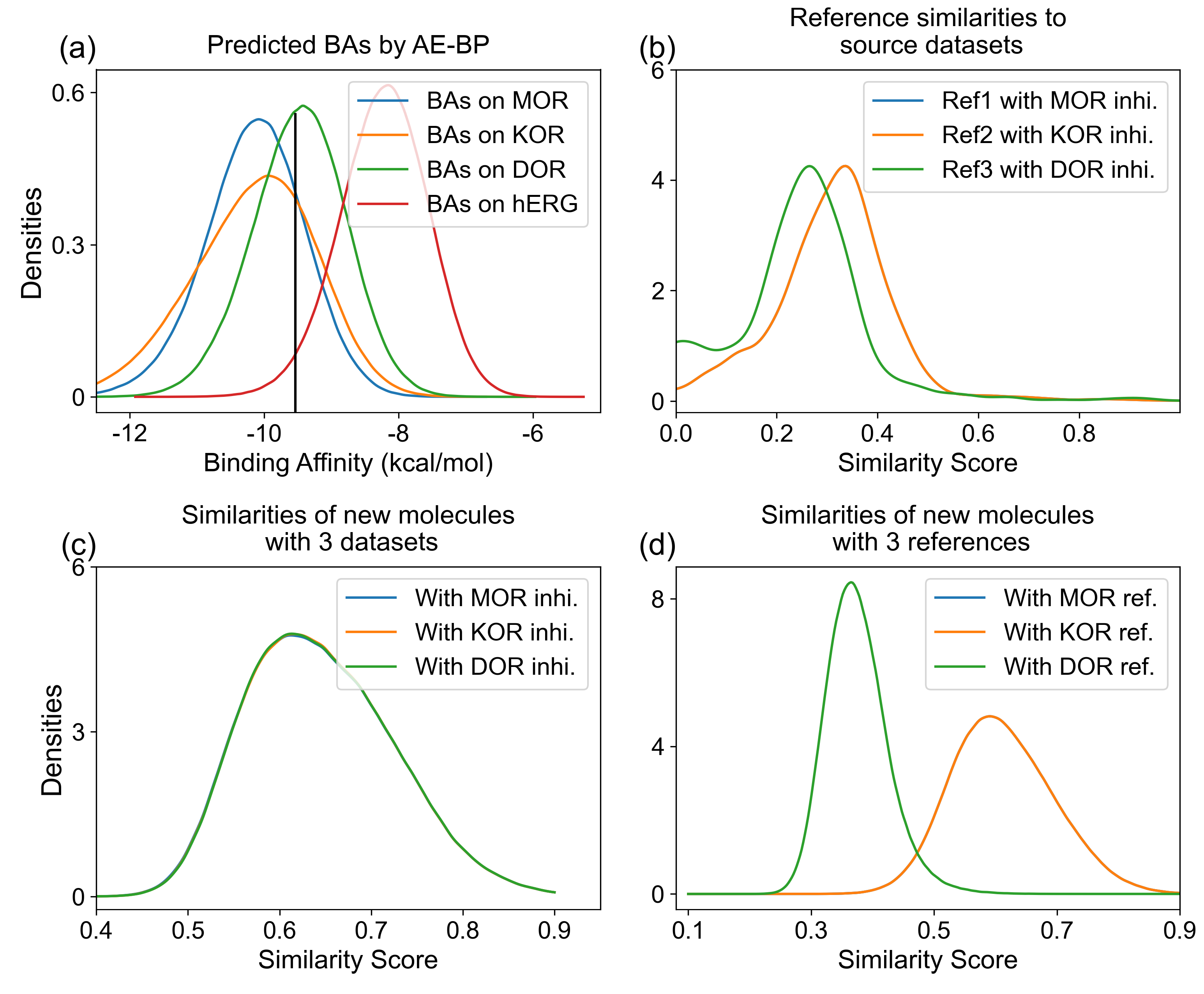} 
			  	\caption{{\footnotesize The BA distribution of the generated compounds, as well as the similarity score distributions regarding two reference compounds, training data, and generated compounds. a. BA distributions of the generated molecules predicted by AE-BPs. The unit for BA is kcal/mol. b. The distributions of similarity scores between reference compounds and corresponding source inhibitor dataset. c. The distributions of similarity scores between generated compounds and three inhibitor datasets. d. The distributions of similarity scores between generated compounds and three reference compounds.
			  	} }
			  	\label{Fig:distributions-BA-similarity-ex2}
			  \end{figure} 
			  
			  % 1829616/2119705= 0.8631
			 We conducted a test using two reference compounds to generate novel molecules, aiming to improve the molecular similarities to the training data and enhance the accuracy of  BA  predictions. The same two compounds, ChEMBL2048770 and ChEMBL494462, used in the previous experiment were utilized. ChEMBL2048770 exhibits high potency as an inhibitor for both MOR and KOR, with binding affinities of -11.51 and -11.78 kcal/mol, respectively. ChEMBL494462 specifically binds to DOR with a BA value of -11.92 kcal/mol. In this test, ChEMBL2048770 was selected as the reference compound for both MOR and KOR, while ChEMBL494462 served as the reference compound for DOR. We continued to use ChEMBL243195 as the seed compound. The weight coefficient $(\alpha_1,\alpha_2,\alpha_3)$ is set to be $(0.35,0.35,0.3)$ in Eqn \ref{eqn:Langevin}. We utilized our GNC to generate millions of new valid molecules, and these new molecules have an reconstruction ratio of $86.31\%$ through the autoencoder encoding-decoding network.

			 Figure \ref{Fig:distributions-BA-similarity-ex2} presents the results of our molecule generation in the second experiment. Similar to the previous experiment, our GNC successfully generated a significant number of active molecules for MOR, KOR, and DOR targets, while exhibiting weak hERG side effects, as depicted in Figure \ref{Fig:distributions-BA-similarity-ex2}a. As anticipated, the generated molecules showed improved similarity scores with the training data, as observed in Figure \ref{Fig:distributions-BA-similarity-ex2}c. The average similarity score of approximately 0.6 was higher than the score of around 0.45 obtained using three reference compounds in our previous experiment, as shown in Figure \ref{Fig:distributions-BA-similarity}d. Furthermore, in Figures \ref{Fig:distributions-BA-similarity-ex2}d, we can observe that the similarity scores for the MOR and KOR reference compounds are higher than those for the DOR reference compound, which aligns with Figure \ref{Fig:distributions-BA-similarity-ex2}b. It is important to note that the distribution curves of the MOR and KOR reference compounds overlap in both Figure \ref{Fig:distributions-BA-similarity-ex2}b and Figure \ref{Fig:distributions-BA-similarity-ex2}d, as ChEMBL2048770 was used as the reference compound for both MOR and KOR targets.
			   
			 By comparing the two experiments, we can see that utilization of two reference compounds can be an effective approach to design novel molecules yet with high similarity scores to the training data. This, in turn, enables more accurate BA predictions.

			  \subsubsection{Binding affinity reevaluation  with consensus models}

				We use AE-BPs for an initial screening of the generated molecules. By applying  BA  constraints, we generate a large pool of novel molecules that may function as effective multi-target inhibitors. To refine this pool, we employ our consensus models to reevaluate the BAs of the filtered generated molecules. The BA constraints are applied once again by the consensus reevaluations, resulting in a reduced number of compounds for ADMET analysis. Below, we continue to carry out further investigations on the above experiment using two reference compounds.

			\subsection{ADMET analysis}
			
				\begin{table}
				\centering
				\begin{tabular}{c|c}		
					\hline
					Property & Optimal range   \\ 	
					\hline
					FDAMDD & Excellent: 0-0.3; medium: 0.3-0.7; poor: 0.7-1.0  \\
					$\rm F_{20\%}$ & Excellent: 0-0.3; medium: 0.3-0.7; poor: 0.7-1.0  \\
					Log P & The proper range: 0-3 log mol/L \\
					Log S & The proper range: -4-0.5 log mol/L \\
					$\rm T_{1/2}$ & Excellent: 0-0.3; medium: 0.3-0.7; poor: 0.7-1.0  \\
					Caco-2 & The proper range: $>$-5.15 \\
					SAS & The proper range: $<$6 \\
					\hline
				\end{tabular}
				\caption{The optimal ranges of six selected ADMET properties and synthesizability (SAS) used to screen nearly optimal compounds.}
				\label{tab:property-optimal}
			\end{table}

			ADMET (absorption, distribution, metabolism, excretion, and toxicity) plays a critical role in drug discovery and development as it encompasses various attributes related to a compound's pharmacokinetic studies. A promising drug candidate must demonstrate both efficacy on the therapeutic target and compliance with essential ADMET properties. Accurate ADMET predictions are vital in drug design as they enable the screening of new compounds' properties and help mitigate the risk of late-stage attrition.
			
			For systematic ADMET screening, we examined six indexes: FDAMDD, T${1/2}$ and F${20\%}$, Log P, Log S, and Caco-2. To assess these ADMET properties, we utilized the machine-learning predictions provided by ADMETlab 2.0 solvers (https://admetmesh.scbdd.com/) \cite{admetlab,admetlab2}. The provided documentation offers optimal ranges for various ADMET properties. Furthermore, we evaluated the synthetic accessibility score (SAS) of the compounds, employing RDKit for the evaluation. Table \ref{tab:property-optimal} provides the optimal ranges for ADMET properties and SAS. By systematically evaluating the binding effects, ADMET properties, and SAS, we conducted a search for potential compound leads.

			  \subsubsection{Potential optimal drug candidates}

			  	\begin{figure}[ht]
			  	\centering
			  	\includegraphics[width=0.78\linewidth]{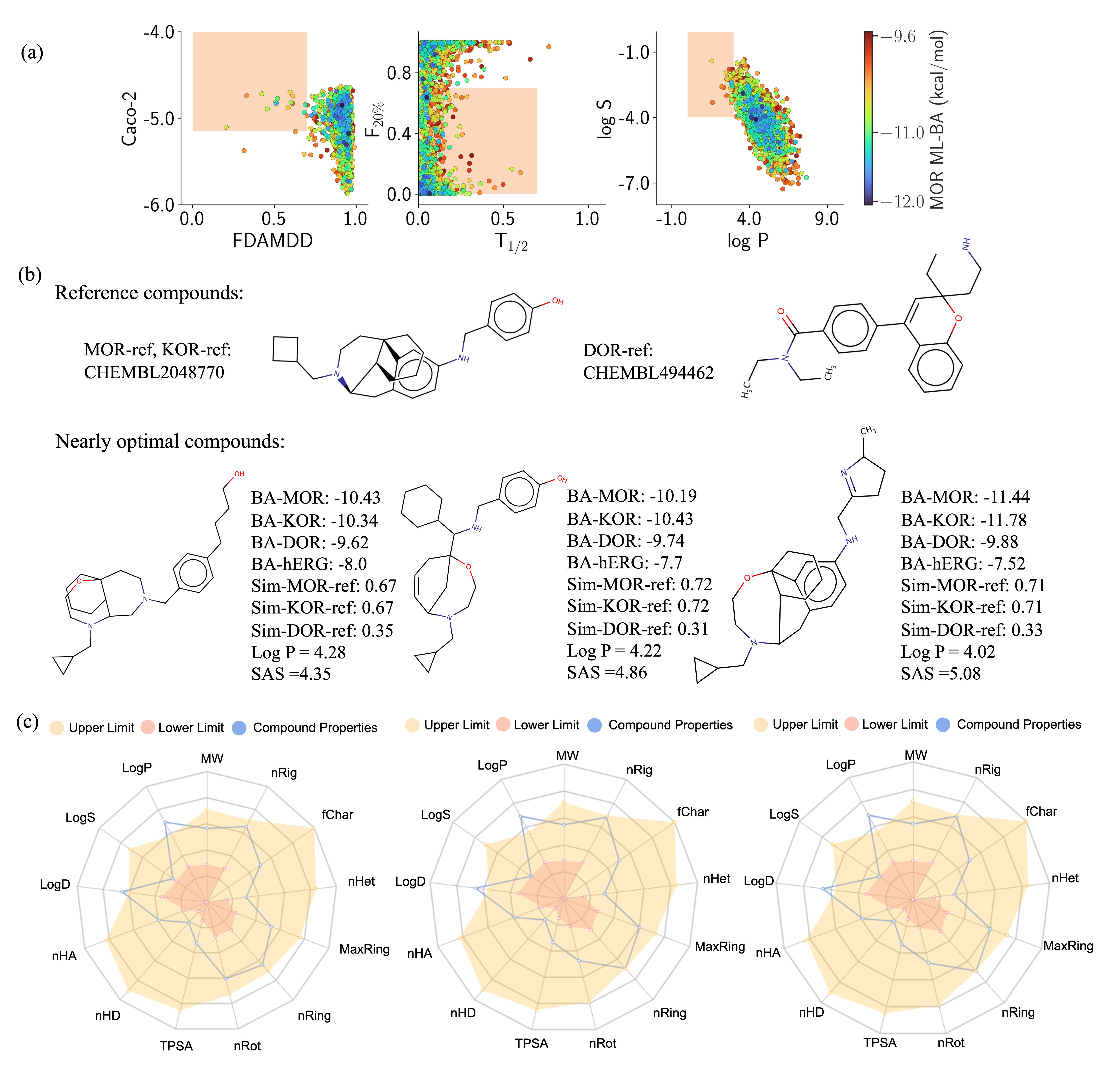} 
			  	\caption{{\footnotesize Identifying nearly optimal compounds. a. ADMET screening of the multi-target molecules. b. Three nearly optimal compounds that satisfy the BA constraint, SAS, and ADMET properties. Their BAs and similarity scores with the two references are presented. c. Additional ADMET prediction from ADMETlab2. More abbreviations: 
			  			MW (Molecular Weight), 
			  			$\log$P (log of octanol/water partition coefficient),  
			  			$\log$S (log of the aqueous  solubility), 
			  			$\log$D (logP at physiological pH 7.4), 
			  			nHA (Number of hydrogen bond acceptors), 
			  			nHD (Number of hydrogen bond donors), 
			  			TPSA (Topological polar surface area), 
			  			nRot (Number of rotatable bonds), 
			  			nRing (Number of rings), 
			  			MaxRing (Number of atoms in the biggest ring), 
			  			nHet (Number of heteroatoms), 
			  			fChar (Formal charge), and 
			  			nRig (Number of rigid bonds). The optimal ranges of these indexes  are shown in table S4 in the Supporting information.
			  	} }
			  	\label{Fig:example-optimal}
			  \end{figure}

		  After performing consensus BA and SAS predictions, we proceeded to evaluate the ADMET properties of the identified multi-target active compounds in the second experiment. Among those, 2155 compounds satisfy the BA constraints and are in the SAS proper range. According to the ADMETlab2 predictions, only a very limited number of compounds can satisfy the various properties. Figure \ref{Fig:example-optimal}a presents the ADMET screening results. The orange frames outline the proper domains for a pair of properties. The color points represent the predicted BA values of the generated compounds on MOR. 
		
		 The first plot in Figure \ref{Fig:example-optimal}a show the distributions of FDA maximum recommended daily dose (FDAMDDs), an index of potential for toxicity, and Caco-2, cell permeability of compounds. The FDAMDD screening reveals that a very small fraction of our generated compounds can satisfy toxicity requirement, despite that nearly half of them can pass the Caco-2 screening. This suggests the necessity of ADMET screening before a new compound is synthesized.
		 
		The second plot in Figure \ref{Fig:example-optimal}a displays the screening results based on two parameters: F$_{20\%}$ (bioavailability of 20\%) and T$_{1/2}$ (half-life). The plot shows that a significant number of the compounds fall within the optimal domain for both indexes. Especially, almost all these generated potent inhibitor can pass the T$_{1/2}$ screening.

		 The third plot in Figure \ref{Fig:example-optimal}a illustrates the screening based on Log P and Log S, which are parameters that relate to the distribution of drugs in the human body. The outline optimal domain covers only a small portion of the plot. While nearly half of the compounds fall within the acceptable range of Log S, very few of these potent inhibitors are within the suggested range for Log P. This suggests that a significant amount of resources are being wasted in early studies.
		 
		The screenings for the compound on FDAMDD and Log P indexes impose strict filtering criteria for finding optimal compounds from the 2155 compounds, which limits the number of potential drug candidates. None of the 2155 potent compounds passed the ADMET screening. To increase the pool of possible drug candidates, we relaxed the Log P requirement by adjusting the proper range to 0-5 log mol/L, as Lipinski's rule suggests Log P less than 5 Log mol/L for an orally active drug \cite{lipinski2012experimental}. Using this new range, we were able to identify three drug candidates, as shown in Figure \ref{Fig:example-optimal}b. The predicted binding affinity values for the four critical targets, as well as their similarity scores to two reference compounds, are provided.

		The ADMETlab2 server was used to evaluate a range of other ADMET indexes for the three nearly optimal compounds. Figure \ref{Fig:example-optimal}c shows that the additional physicochemical properties of the two molecules were within the appropriate ranges, except for Log P and Log D. Log D, which is associated with Log P, refers to the logarithmic value of Log P at physiological pH 7.4. Upon structural optimization of the three molecules, it is possible to achieve simultaneous optimization on both Log P and Log D.

		\subsection{Molecular optimization}
		
		As noted earlier, the FDAMDD and Log P profiles pose obstacles that prevent the generated molecules from becoming optimal drug candidates in the second experiment. To expand the pool of optimal drug candidates, we consider optimizing the generated compounds that exhibit desired BA values and nearly satisfactory ADMET properties. Log P index is the objective we strive to optimize.
		
		\subsubsection{Log P optimization}
		
		 The polarity of a molecule can influence its Log P value. A highly polar molecules may have lower Log P values, as they are more soluble in the aqueous phase and less likely to partition into the lipid phase. The relationship between molecular polarization and Log P can be complex and depends on other factors such as molecular size, shape, and functional groups. To induce more polarization, we propose replacing a hydrogen atom with a hydroxyl group. These modified molecules are then subjected to scrutiny for molecular validity through our encoding-decoding process. The reconstructed molecules are subsequently evaluated for their BAs and ADMET properties using our consensus BA-predictors and ADMETlab2. This optimization process may yield compounds that are closer to being optimal in terms of desired BAs and ADMET profiles. We tested this strategy on the three aforementioned molecules, which were nearly optimal except for their Log P profiles.
		 
		 	\begin{figure}[ht]
		 	\centering
		 	\includegraphics[width=0.9\linewidth]{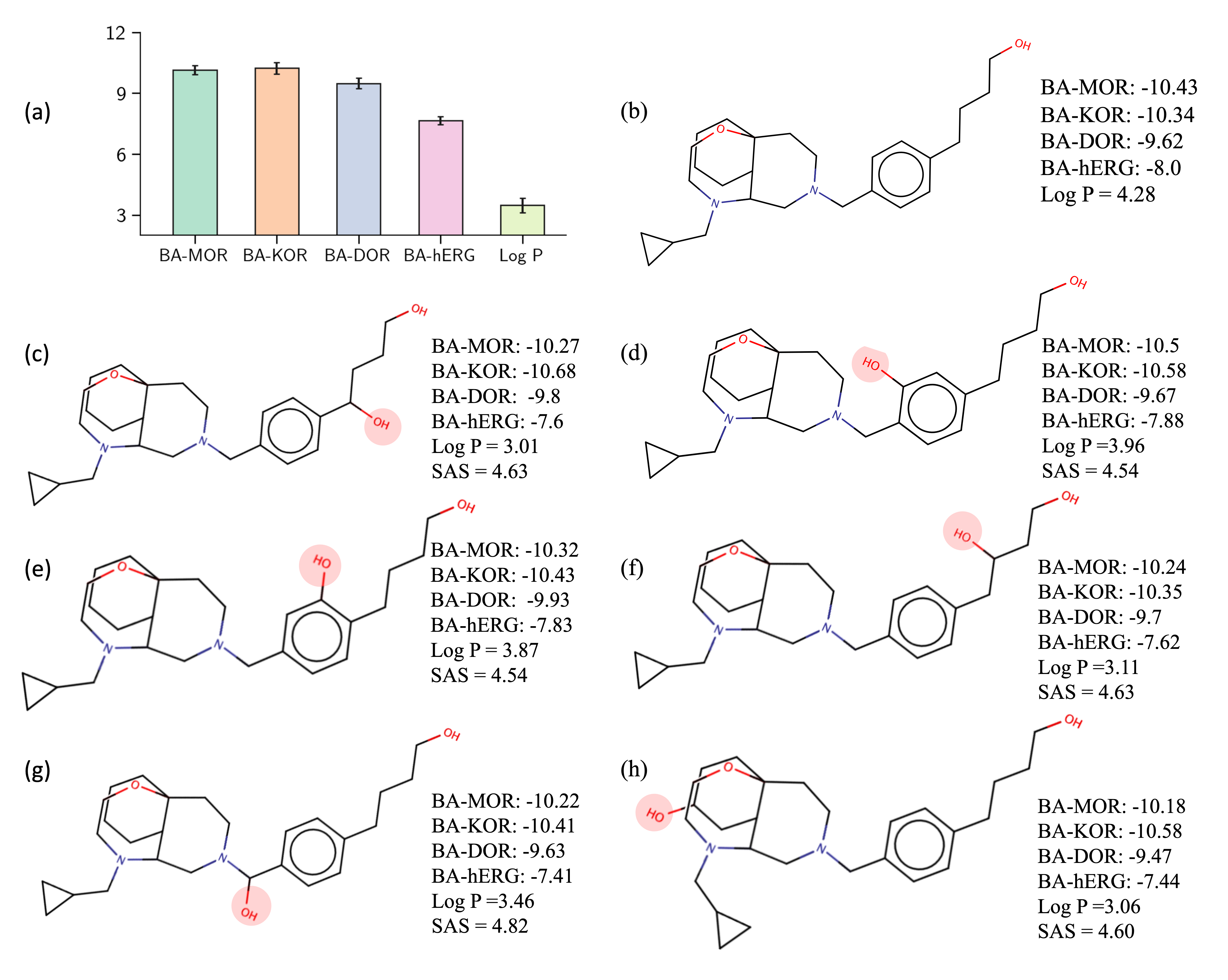} 
		 	\caption{{\footnotesize Results of Log P optimization for one nearly optimal compound. The y-axis indicates the magnitude of the BA and Log P value. The BA values are all negative and the Log P values are all positive. a: the statistics of the predicted BA values on the four critical targets and the predicted log P values of the induced 22 compounds. b: the original generated compound from our GNC that are considered to be nearly optimal compound.  c-h: six of the 22 derived compound that have high BAs and improved Log P values. 
		 	} }
		 	\label{Fig:optimization-logP-case1}
		 \end{figure} 
		 
		 Figure \ref{Fig:optimization-logP-case1} displays the molecular optimization on its Log P profile for the first nearly optimal compound in Figure \ref{Fig:example-optimal}b. By replacing one hydrogen atom with a hydroxyl group on the molecule, we obtained 22 new valid molecules. Figure \ref{Fig:optimization-logP-case1}a shows the statistics of the 22 new compounds on the magnitude of predicted BA values on MOR, KOR, DOR, and hERG, as well as their Log P values. The original compound with its BAs and Log P value is shown in Figure \ref{Fig:optimization-logP-case1}b. The average predicted BA values on MOR, KOR, DOR, and hERG are -10.15, -10.24, -9.49, and -7.65 kcal/mol. The original generated compound has predicted BA values of -10.43, -10.34, -9.62, and -8.0 kcal/mol, as well as Log P of 4.28. These 22 derived compounds exhibited slightly reduced potency on the four critical targets but improved Log P profiles. The optimization process alleviated hERG side effect and reduced Log P values.  The Figure \ref{Fig:optimization-logP-case1}c-h displays the six derived molecules with highest average BA potency on the MOR, KOR, and DOR. The red circles highlight the positions where a hydrogen atom is replaced. All the six compounds showed improved Log P profiles. In addition, five of them were all predicted to be effective inhibitors on MOR, KOR, and DOR without hERG side effects. Among the 22 derived compounds, 9 were predicted to be effective on the three targets while their Log P values are less than 4.0. These results demonstrate the effectiveness of the optimization process in offering drug candidates with improved pharmacokinetic profiles. In addition to the potency and Log P, other pharmacokinetics properties are taken into account again to identify nearly optimal compounds. Unfortunately, none of the 22 compounds met the criteria for binding affinity and ADMET properties, thereby ruling out their candidacy as new drugs. This reflects the challenge in molecular optimization, where multi-objective optimization is characterized by the inherent trade-off between improving one property at the expense of another.
		 
		 \begin{figure}[ht]
		 	\centering
		 	\includegraphics[width=0.9\linewidth]{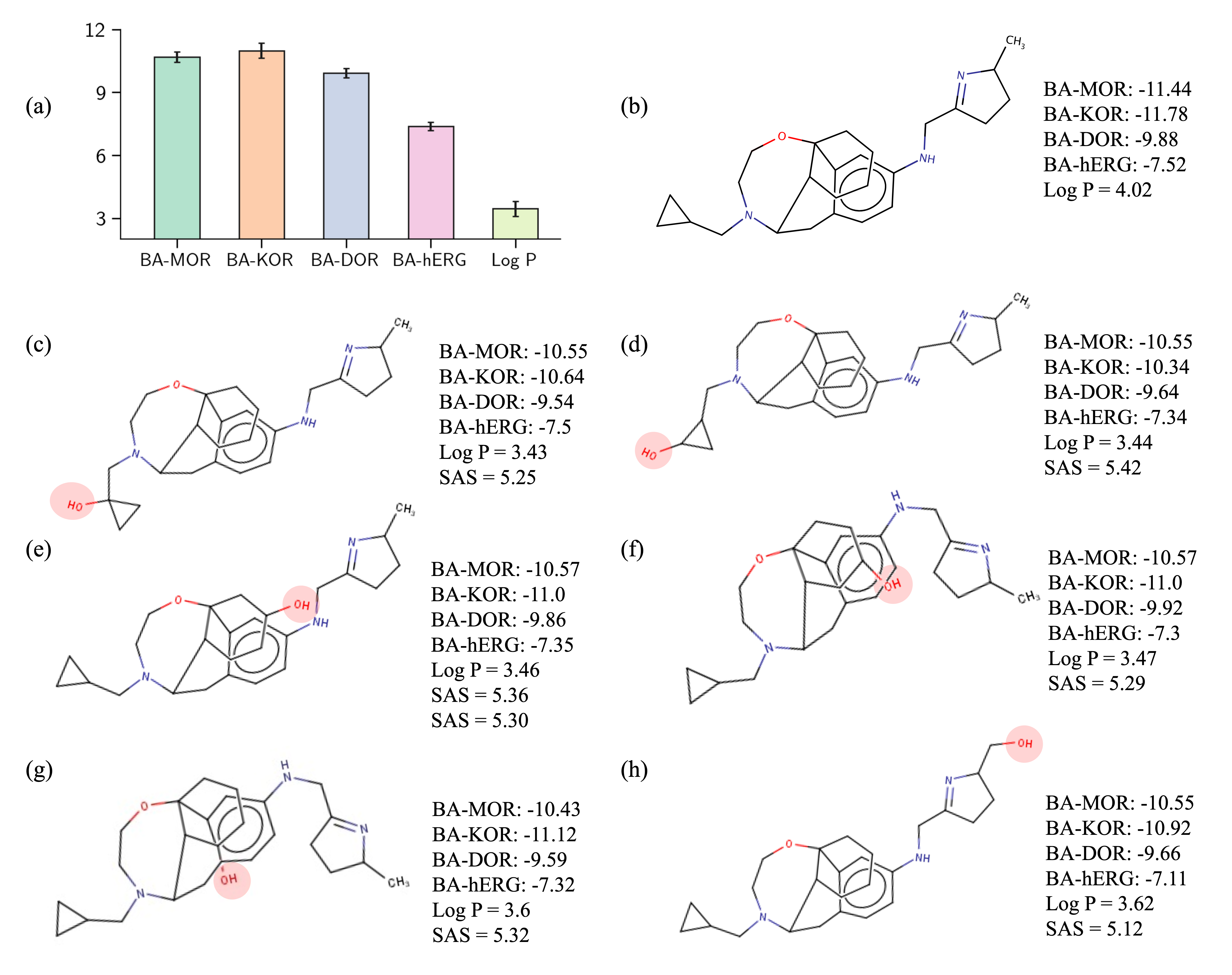} 
		 	\caption{{\footnotesize Results of Log P optimization for one nearly optimal compound. a: the statistics of the predicted BA values on the four critical targets and the predicted log P values of the derived 22 compounds. The y-axis indicates the magnitude of the BA and Log P value. The BA values are all negative and the Log P values are all positive. b: the original generated compound from our GNC that are considered to be nearly optimal compound.  c-h: six of the 22 derived compound that have high BAs and improved Log P values. 
		 	} }
		 	\label{Fig:optimization-logP-case2}
		 \end{figure}

		 We applied the Log P optimization approach to the other two nearly optimal compounds shown in Figure \ref{Fig:optimization-logP-case1}b and screened them for potency and ADMET properties. Upon replacing a hydrogen atom with a hydroxyl group, we obtained 20 new valid molecules for the second compound and 21 new valid molecules for the third compound. Among these, 11 inhibitors effectively targeted multiple receptors (MOR, KOR, and DOR) for the second compound, while the third compound yielded 20 such inhibitors. The average predicted BA values for the 20 new compounds on MOR, KOR, DOR, and hERG were -9.98, -10.06, -9.62, and -7.52 kcal/mol, respectively. Similarly, for the 21 new compounds, the average predicted BA values were -10.68, -10.99, -9.92, and -7.38 kcal/mol, respectively. The average Log P values for the 20 compounds were 3.47, and for the 21 compounds, it was 3.85. Among the 20 compounds, 11 were effective on all three opioid receptors, and 18 had Log P values less than 4. Ten compounds exhibited desired BA and Log P values. Among the 21 compounds, 20 were effective on all three opioid receptors, and 14 had Log P values less than 4. Thirteen compounds showed desired BA and Log P values. Figure \ref{Fig:optimization-logP-case2} illustrates some of our optimization results for the third compound in Figure \ref{Fig:optimization-logP-case1}b. The six best derived compounds, in terms of Log P value, are presented. It can be observed that modifications to various functional groups of the molecule contributed to a reduction in Log P values.

\begin{figure}[ht]
		 	\centering
		 	\includegraphics[width=0.85\linewidth]{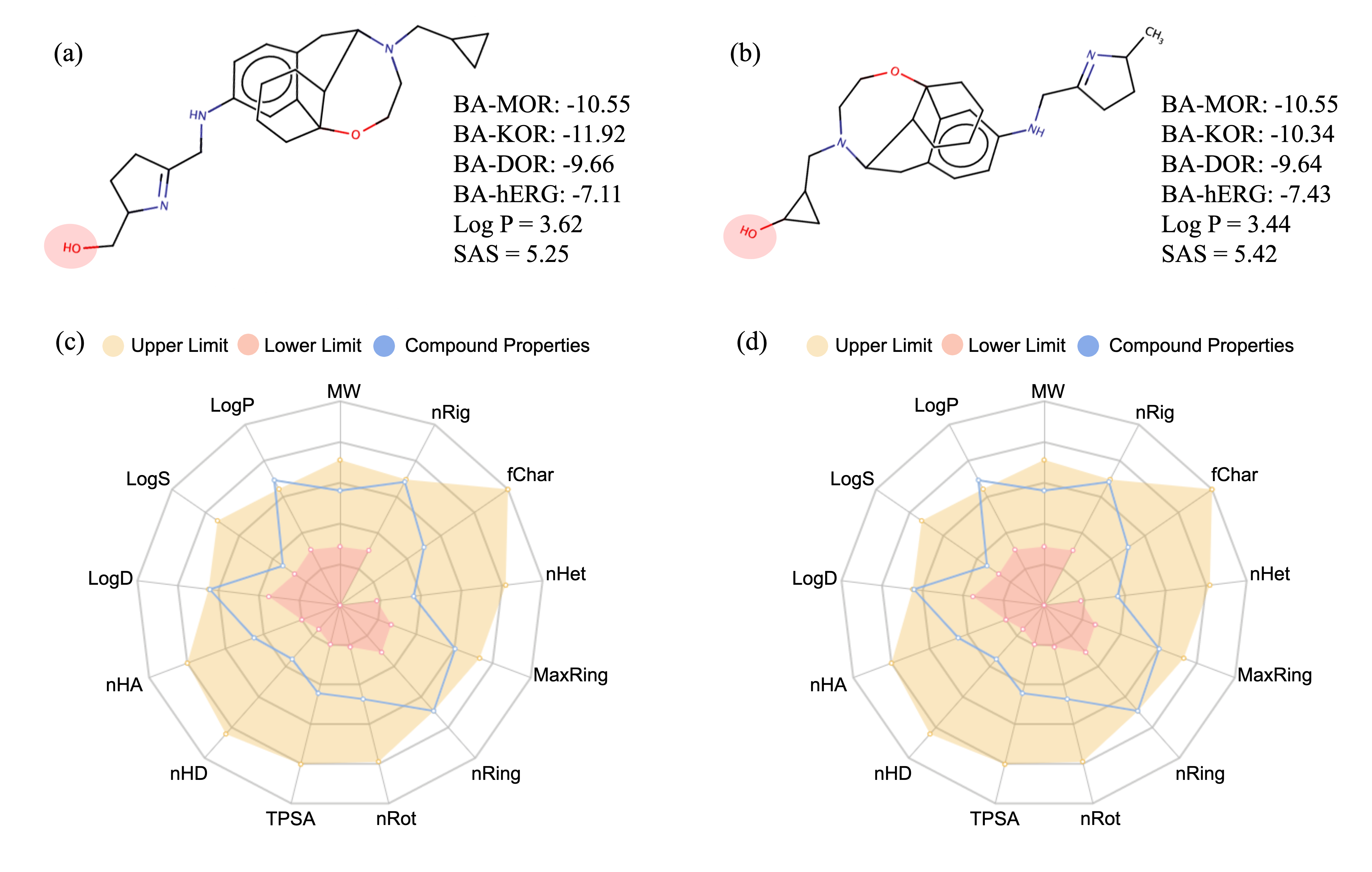} 
		 	\caption{{\footnotesize Two nearly optimal compounds derived from the Log P optimization on one given nearly optimal compound. 
		 	} }
		 	\label{Fig: two-logp-optimization}
		 \end{figure}

		Through molecular optimization of the three nearly optimal compounds, we have obtained some new compounds. Among these compounds, a significant number exhibit desired BA and Log P values, thereby providing more promising candidates as nearly optimal compounds. By screening these compounds on their BAs and additional ADMET properties using our BA predictors and ADMETlab2, we have identified two compounds with improved Log P profiles that meet all BA and ADMET requirements. These two compounds, derived from applying Log P optimization to the third nearly optimal compound shown in Figure \ref{Fig:example-optimal}b, are presented in Figure \ref{Fig: two-logp-optimization}. The positions where hydrogen atoms were replaced are highlighted in Figure \ref{Fig: two-logp-optimization}a and b. The screening results for a series of physicochemical properties are depicted in Figure \ref{Fig: two-logp-optimization}c and d. In comparison to the aforementioned three nearly optimal compounds, the two derived compounds exhibit improved Log P and Log D profiles.

		The optimization process is beneficial to the generation of additional drug candidates in an efficient way. We screened millions of compounds before we could identify three nearly optimal compounds in Figure \ref{Fig:example-optimal}b. The Log P optimizations of the three compounds provide the two additional compounds with improved Log P profiles in a short time.

		\subsubsection{Molecular interactions between opioid receptors and effective inhibitors}

\begin{figure}[ht]
				\centering
				\includegraphics[width=0.9\linewidth]{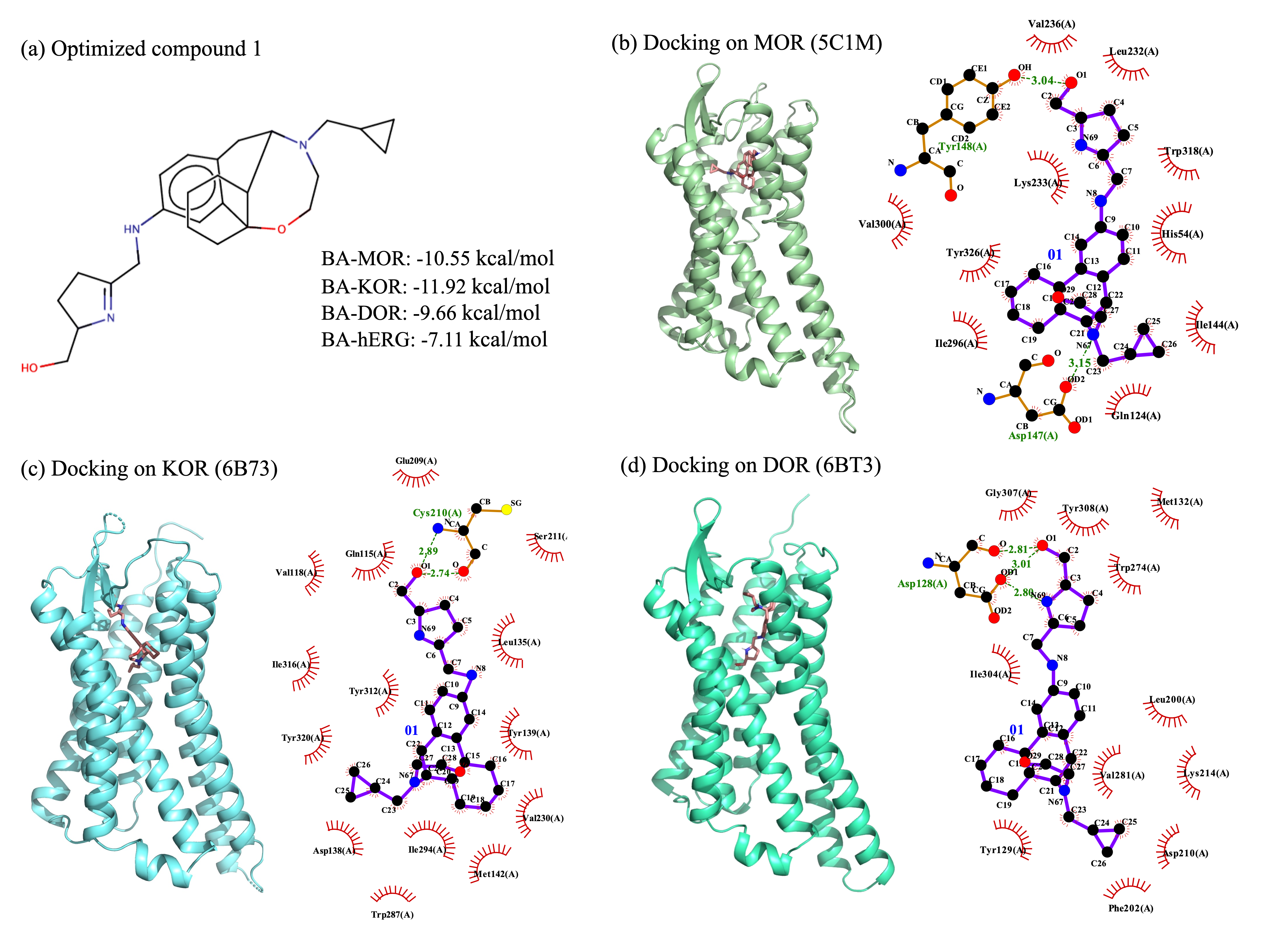} 
				\caption{{\footnotesize The molecular ducking poses and interactions of one optimized compound with three opioid receptors. The PDB IDs for the three receptors are shown in parenthesis.
				} }
				\label{Fig: interaction-optimized-1}
			\end{figure}

	   It is crucial to understand the molecular mechanism of drug-target interactions in identifying desired drug candidates. To predict the docking poses  of one drug candidate to opioid receptors, namely, MOR, KOR, and DOR, we utilized the molecular docking software AutoDock Vina \cite{trott2010autodock}. The three receptors are in the same protein family and share high structural similarities. Above we identified five nearly optimal drug candidates including two derived compounds. The compound in Figure \ref{Fig: two-logp-optimization}a showed high potency profiles and improved ADMET properties. Its molecular docking poses on the receptors are depicted in Figure \ref{Fig: interaction-optimized-1}. 
			  
		It is observed that hydrogen bonds play critical roles in the molecular interactions. There are at least two hydrogen bonds in each drug-target interaction system, which contribute to the high potency of the molecules on the receptors. The molecule in Figure \ref{Fig: two-logp-optimization}a was derived with the Log P optimization by replacing a hydrogen atom in methyl group with a hydroxyl group. The hydroxyl group itself plays a critical role in molecular interactions with the three receptors by forming hydrogen bonds. In its interaction with MOR, a hydrogen bond is formed between one oxygen atom on residual Tyr148(A) of MOR and the oxygen in the hydroxyl group. The second hydrogen bond in this interaction occurs between one nitrogen atom of the molecule and an oxygen atom in residual Asp147(A) of MOR. When interacting with KOR, the oxygen atom in the hydroxyl group forms hydrogen bonds with a nitrogen atom and an oxygen atom on residual Cys210(A), respectively. In its interaction with DOR, two hydrogen bonds are formed between the hydroxyl group and two oxygen atoms on residual Asp128(A) of DOR. Additionally, another hydrogen bond exists between a hydrogen atom on Asp128(A) and a nitrogen atom on the compound.
		
		\subsection{Additional nearly optimal compounds}
		
		Above, we have presented a comprehensive overview of the workflow within our complex generative network for the discovery of novel multi-target compounds. These compounds exhibit drug-like potential based on machine learning predictions and hold promise for  OUD treatment. Utilizing our  GNC, we generated additional multi-target compounds using different reference and seed compounds. Through screening their BA values, SAS scores, and ADMET properties, we identified additional compounds with drug-like potential for treating OUD. Further details on these compounds can be found in the Supporting Information.
		
		\section{Discussion}
		
		\subsection{Designing analog drugs of the approved medications}
			
		Currently, the US  FDA  has approved three medications, namely methadone, buprenorphine, and naltrexone, for the treatment of  OUD. These medications exert their pharmacological effects by targeting MOR, KOR, and DOR. Additionally, naloxone is a crucial medication used for the treatment of opioid overdose. We are interested in generating potential analogs of these four medications. We utilize these medications as reference compounds. Our focus is on designing analogs that exhibit simultaneous activity on MOR, KOR, and DOR. To achieve this, we employ both molecular generation and optimization approaches.  
		
		\subsubsection{Designing buprenorphine analogs}

		 Buprenorphine acts as a partial agonist for the MOR receptor and an antagonist for the KOR receptor. It can alleviate opioid withdrawal symptoms, reduce the effects of injected opioids, and provide protection against overdose \cite{bell2014pharmacological}. Buprenorphine has a ceiling effect on euphoria and carries a lower risk of respiratory depression compared to methadone \cite{mattick2014buprenorphine}. 
				
\begin{figure}[ht]
			\centering
			\includegraphics[width=0.85\linewidth]{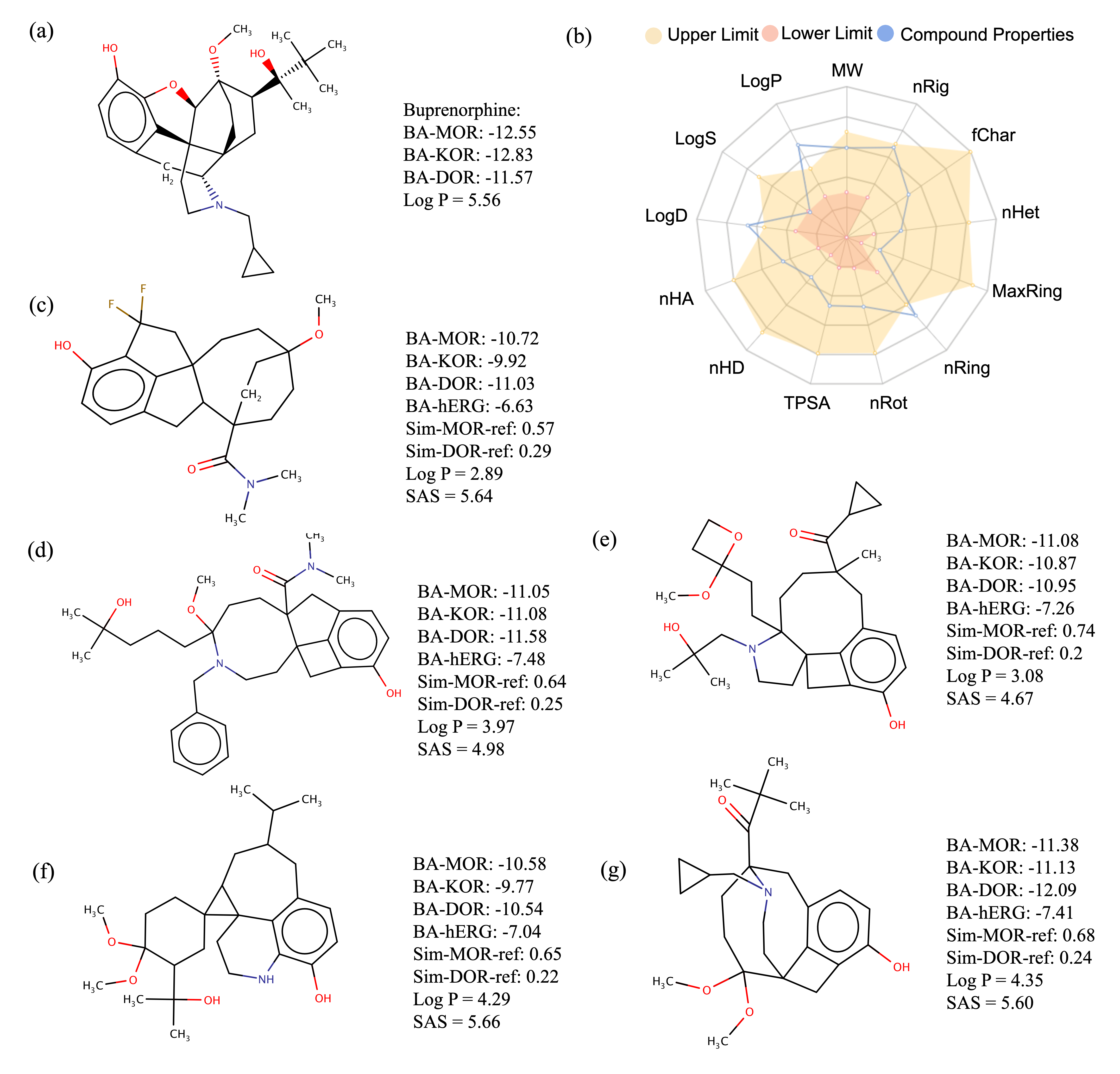} 
			\caption{{\footnotesize a: The 2D structure of buprenorphine. b-h: Several nearly optimal buprenorphine analogs that could be effective on MOR, KOR, and DOR. The predicted BA values on the three receptors and hERG, the similarity scores to reference compounds, as well as their predicted Log P values, are provided.
			} }
			\label{Fig: analogs-buprenorphine}
		\end{figure}

		Buprenorphine exhibits activity on MOR, KOR, and DOR with BA values of -12.55, -12.83, and -11.57 kcal/mol, respectively. To generate analogs of buprenorphine, we use it as the reference compound for both MOR and KOR. For DOR, we utilize ChEMBL494462 as the reference compound, which has a BA value of -11.92 kcal/mol. In applying the molecular generator, a weight coefficient of 0.8 is assigned to the latent vectors of buprenorphine, while the AE latent vector of ChEMBL494462 is given a weight coefficient of 0.2. This weighting scheme increases the likelihood of generating buprenorphine analogs, as the higher weight assigned to buprenorphine promotes a greater resemblance to this reference compound. Our GNC generated millions of novel molecules. From this vast pool, we identified five compounds that were nearly optimal, each possessing Log P values below 5. The synthetic accessibility score (SAS) less than 6 is the suggested proper range. A lower SAS indicates a higher level of ease in synthesizing the compound. The five compounds have a SAS of less than 5.7, with two scores below 4.  Therefore, these five compounds can be synthesized relatively easily. These highly potent compounds on all three receptors are depicted in Figure \ref{Fig: analogs-buprenorphine}. 
		
		Buprenorphine, with BA values being close to -13 kcal/mol on the three receptors, resides at the outer bounds of the BA distributions in the training data. In Figure \ref{Fig: analogs-buprenorphine}, the identified compounds exhibit predicted BAs around -11 kcal/mol for the receptors. Moreover, they exhibit low predicted hERG side effects and Log P values are below 5. Among the compounds, one has a Log P value being below 3, two have Log P values between 3 and 4, and two have Log P values between 4 and 5. In comparison to buprenorphine whose Log P value is 5.56, the five identified compounds possess superior Log P profiles.

		The five generated compounds exhibit similarities ranging from 0.57 to 0.74 with buprenorphine. Since novel molecules are generated with potent reference compounds, they can inherit certain moieties from them, which can facilitate effective binding to the three receptors. These new compounds are observed to preserve some functional groups. For instance, buprenorphine contains methyl groups, a benzene ring, trimethylene, and hydroxyl groups. Many of the generated compounds retain these groups, particularly those with higher similarity scores to buprenorphine. Among the five compounds, the one depicted in Figure \ref{Fig: analogs-buprenorphine}b is the least similar to buprenorphine, with a similarity score of 0.57. However, it introduces two fluorine atoms as new elements while still preserving a benzene ring, several hydroxyl groups, and methyl groups. Notably, it possesses the best Log P profile among the five new compounds. The retention of these functional groups in the five molecules contributes to their binding potency on the receptors.

		We apply our Log P optimization strategy to the three compounds depicted in Figure \ref{Fig: analogs-buprenorphine}f-g, as their Log P values exceed 4. By replacing hydrogen atoms with hydroxyl groups on these three molecules, we generate 14, and 18 new molecules for each respective compound. This optimization approach proves valuable in generating additional nearly optimal compounds based on the molecules in Figure \ref{Fig: analogs-buprenorphine}f and \ref{Fig: analogs-buprenorphine}g. The derived nearly optimal compounds, exhibiting desired binding affinity and ADMET properties, are presented in Figure S2 and Figure S3 in the Supporting Information.

\begin{figure}[ht]
			\centering
			\includegraphics[width=0.9\linewidth]{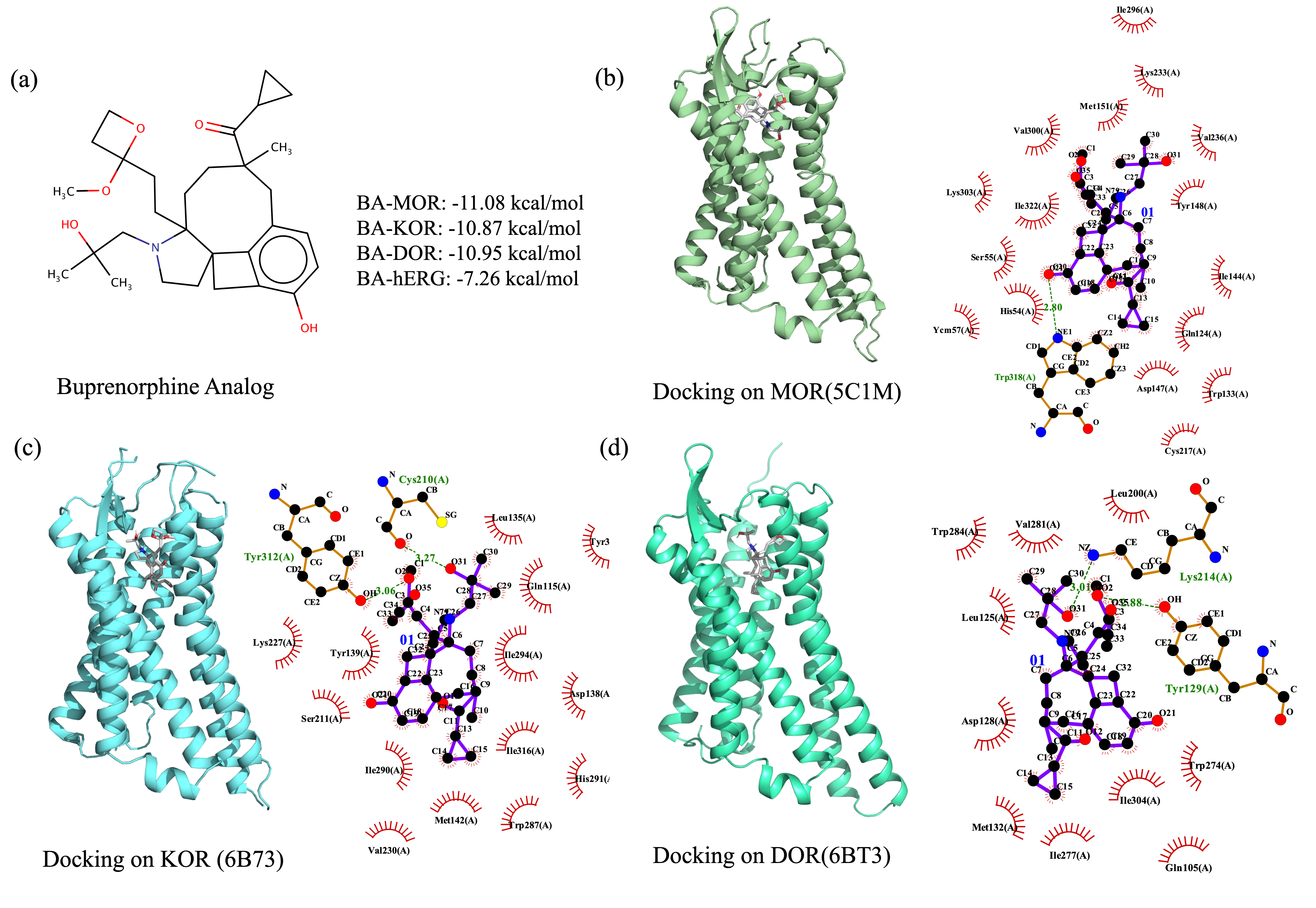} 
			\caption{{\footnotesize The optimization results for buprenorphine analog in Figure 9g in the paper. a: 2D structure of one one nearly optimal buprenorphine analog. b-e: the five derived molecules that have improved Log P profiles. The hydrogen replacement positions were highlighted in red.
			} }
			\label{Fig: analog-buprenorphine-docking-docking}
		\end{figure}

		The compound depicted in Figure \ref{Fig: analogs-buprenorphine}e exhibits the highest similarity to buprenorphine, with a similarity score of 0.74. It demonstrates high potency on all three receptors while exhibiting a low hERG side effect. We also employed Autodock Vina software to predict the molecular interactions with the three receptors. As illustrated in Figure \ref{Fig: analog-buprenorphine-docking-docking}, the compound establishes multiple hydrogen bonds with the receptors. In its interaction with the MOR receptor, a hydrogen bond is formed between the oxygen atom on the benzene ring of the molecule and the nitrogen atom on the residue Trp318(A). Regarding its interaction with the KOR receptor, two hydrogen bonds are formed. One is established between an oxygen atom of the molecule and an oxygen atom on the residue Tyr312(A), while the other is formed between an oxygen atom of the molecule and a sulfur atom on the residue Cys210(A). Additionally, the molecule forms two hydrogen bonds with the DOR receptor. One bond is created between an oxygen atom on the molecule and an oxygen atom on the residue Tyr129(A), while the other bond is formed between an oxygen atom on the molecule and a nitrogen atom on the residue Lys214(A).
		
		\subsubsection{Designing naltrexone analogs}
				
		Naltrexone has BA values of -12.55, -12.11, and -10.48 kcal/mol on MOR, KOR, and DOR, respectively. It functions as an antagonist for both MOR and KOR. Its KOR antagonist properties have been linked to mood improvements in individuals with OUD \cite{weerts2008differences}. However, it faces challenges regarding low adherence among individuals addicted to opioids \cite{morgan2018injectable}.

\begin{figure}[ht]
			\centering
				\includegraphics[width=0.85\linewidth]{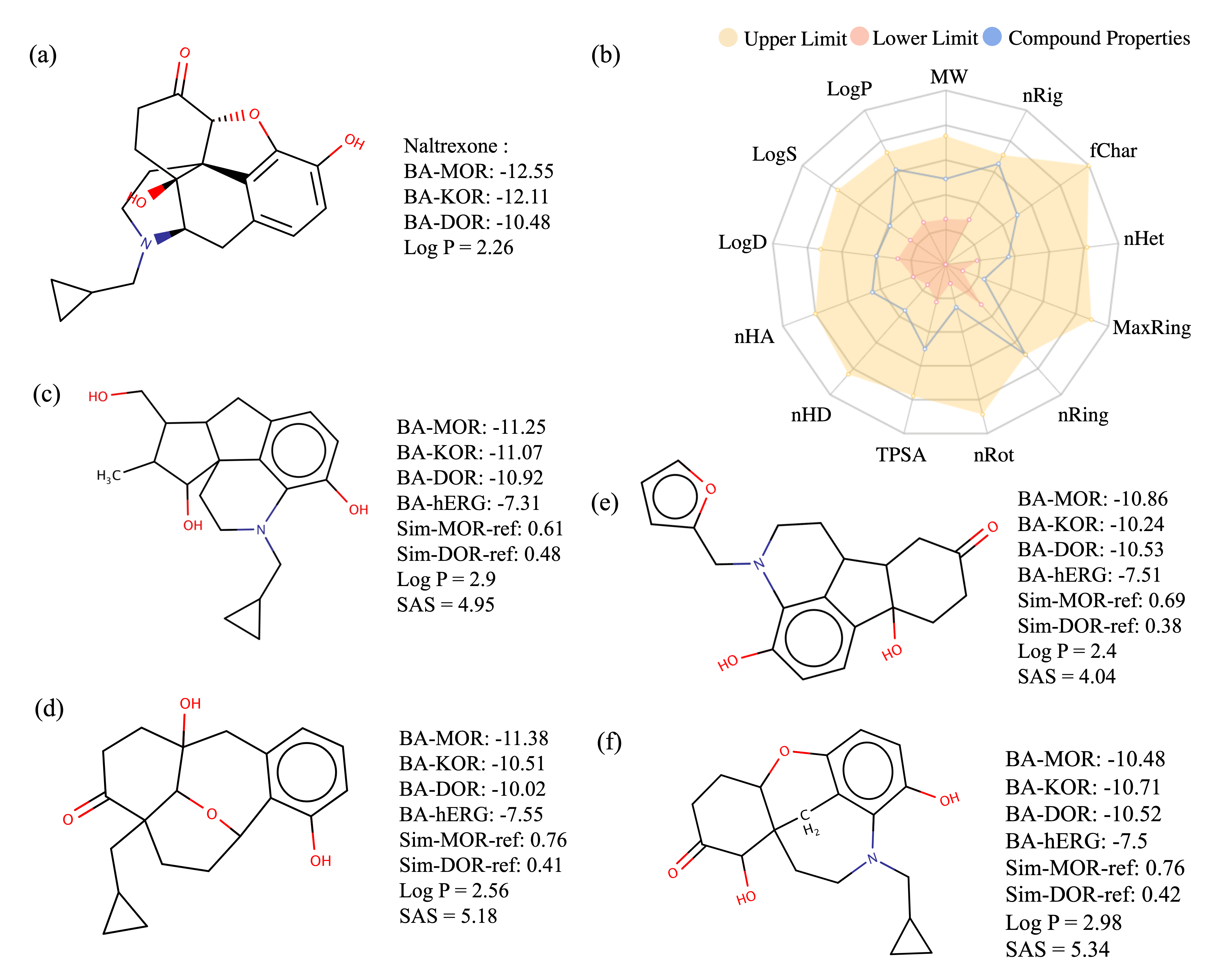} 
					\caption{{\footnotesize a: The 2D structure of naltrexone. b-f: Several nearly optimal naltrexone analogs that could be effective on the multiple receptors. The predicted BA values on the critical targets, the similarity scores to reference compounds, as well as their predicted Log P values, are provided. 
					} }
				\label{Fig: analogs-naltrexone}
		  \end{figure}

		We employed our GNC to generate analogs of naltrexone. Naltrexone was used as the reference compound for MOR and KOR, while ChEMBL56585 served as the reference compound for DOR. ChEMBL56585 displayed BA values of -12.26, -13.64, and -12.35 kcal/mol on MOR, KOR, and DOR, respectively. In applying the molecular generator, weight coefficients of 0.8 and 0.2 are assigned to the latent vectors of naltrexone and ChEMBL56585, respectively. From the millions of generated compounds, we identified four analogs that were nearly optimal, as depicted in Figure \ref{Fig: analogs-naltrexone}. The four compounds have SAS values close to 4 or 5. Therefore, they can be synthesized easily compared to the above buprenophine analogs.
	
		The similarity of these compounds to naltrexone ranged from 0.61 to 0.76. The least similar compound exhibited molecular novelty while still possessing similar functional groups as naltrexone, such as a benzene ring, hydroxyl group, and other rings. Naltrexone displayed an optimal Log P profile with a value of 2.26, and the resulting five analogs inherited this favorable Log P profile. This emphasizes the importance of selecting reference compounds with desired physicochemical properties, eliminating the need for further optimization to improve the Log P profiles of the derived compounds. The naltrexone analog shown in Figure \ref{Fig: analogs-naltrexone}c demonstrates promising binding effects, with a detailed illustration of its molecular interactions provided in Figure S4 in the Supporting Information.
		
	\subsubsection{Designing methadone analogs}

	Methadone functions as a full agonist on the MOR receptor and can alleviate withdrawal and craving symptoms \cite{fareed2010effect}. Its extended half-life and diminished drug-like effects, such as euphoria, result in fewer withdrawal symptoms and reduced potential for reinforcing behavior compared to other opioids \cite{brown2004methadone}. However, methadone carries a risk of respiratory depression in cases of overdose.
	
	Methadone demonstrates BA values of -11.84, -8.99, and -8.54 kcal/mol on MOR, KOR, and DOR, respectively. Designing a multi-target methadone analog presents a challenging task, given that methadone primarily exhibits potency on the MOR receptor. In our molecular generator, methadone serves as the reference compound for MOR and KOR, while other potent compounds were used as the reference for DOR. To effectively generate analogs of methadone, assigning a high weight coefficient to methadone is necessary. However, this approach results in a limited number of compounds that effectively target all three receptors. Additionally, due to the strict ADMET requirements, none of our generated compounds can be considered nearly optimal.

	\subsubsection{Designing naloxone analogs}
	 
	 Naloxone is an opioid antagonist used for reversing respiratory depression in cases of opioid overdose. It exhibits a high affinity that allows it to displace opioid drugs, thereby counteracting their respiratory depression effects. It do not stimulate the opioid receptors and therefore does not cause sedation, analgesia, respiratory depression, and euphoria \cite{mclellan2000drug}. Naloxone specifically exhibits BA values of -11.47, -10.96, and -9.8 kcal/mol on the MOR, KOR, and DOR receptors, respectively.

\begin{figure}[ht]
		\centering
		\includegraphics[width=0.85\linewidth]{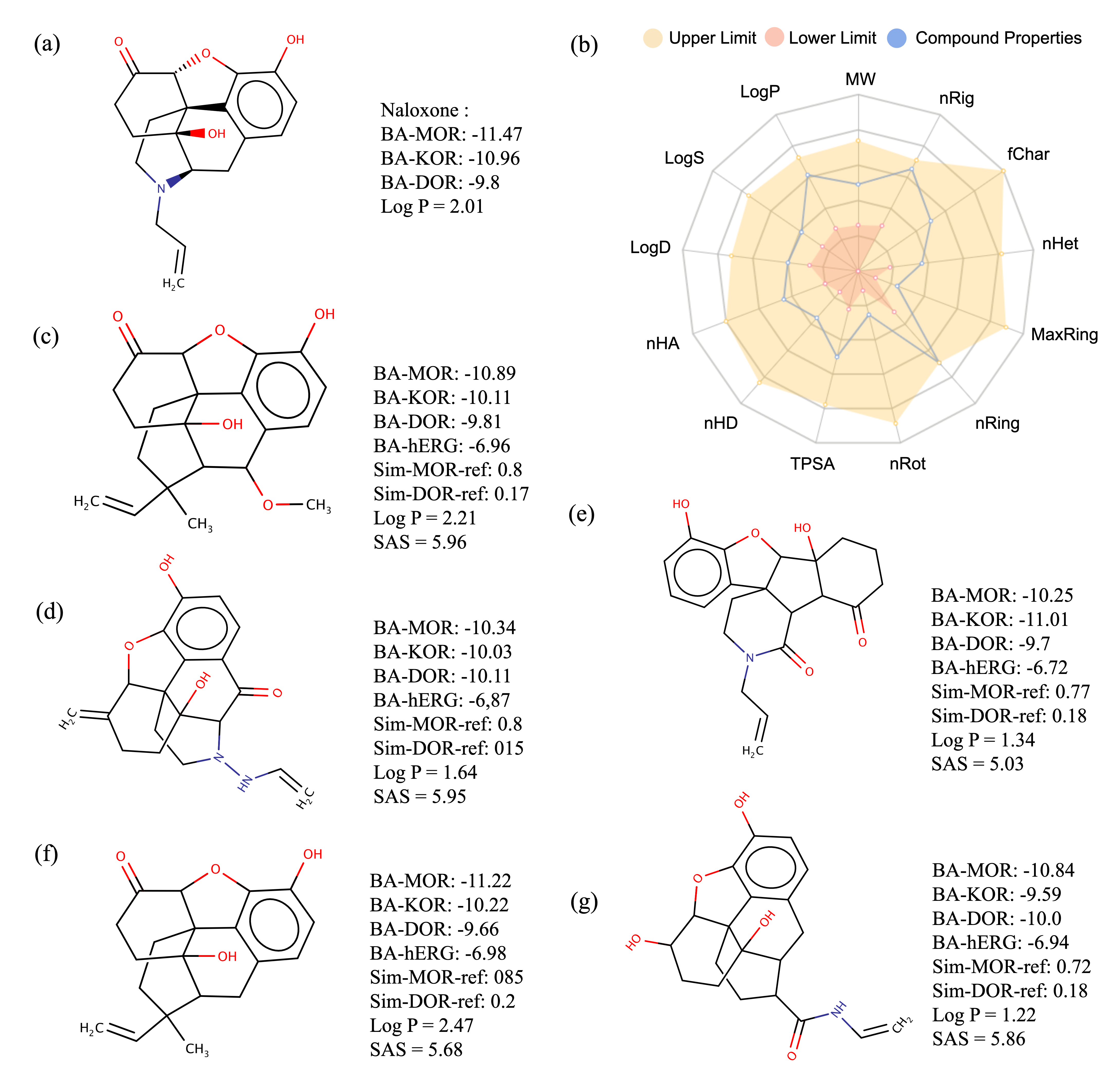} 
		\caption{{\footnotesize a: The 2D structure of naloxone. b-f: Several nearly optimal naloxone analogs that could be effective on the multiple receptors. The predicted BA values on the critical targets, the similarity scores to reference compounds, as well as their predicted Log P values, were provided. 
		} }
		\label{Fig: analogs-naloxone}
	\end{figure}

	 We utilize naloxone as the reference compound for the MOR and KOR receptors, while ChEMBL494462 serves as the reference compound for DOR. Compound ChEMBL494462 demonstrates BA values of -12.26, -13.64, and -12.35 kcal/mol on the MOR, KOR, and DOR receptors, respectively. In applying the molecular generator, weight coefficients of 0.8 and 0.2 are assigned to the latent vectors of naloxone and ChEMBL494462, respectively. From millions of generated compounds, we identified five molecules that exhibit nearly optimal properties, as illustrated in Figure \ref{Fig: analogs-naloxone}. The five compounds, except for compound e, have SAS values close to six. These SAS values indicate a relatively high level of difficulty in synthesizing these compounds.
	 
	  The high weight for naloxone has a significant impact on the structures of its derivatives. As depicted in Figure \ref{Fig: analogs-naloxone},  there are high degree of similarities between these compounds and naloxone. The smallest similarity value observed with naloxone was 0.72, indicating a strong resemblance in terms of molecular structures and functional groups. All five derivatives contain critical pharmacophores such as methyl groups, hydroxyl groups, benzene rings, and other related ring structures, similar to naloxone. Moreover, these analogs possess optimal Log P profiles, akin to the design of naltrexone analogs. This can be attributed to naloxone's favorable physicochemical properties, which facilitate the derivatives' adherence to ADMET requirements. The naloxone analog presented in Figure \ref{Fig: analogs-naloxone}c demonstrates promising binding effects, with a detailed depiction of its molecular interactions provided in Figure S5 in the Supporting Information.

		\subsection{Key factors in designing optimal compounds}

		We proposed a molecular generator based on stochastic differential equation in the latent space of autoencoder networks. Molecules are represented by the AE latent vectors. Multiple reference compounds are used to guide the design of novel molecules such that the generated compounds will inherit similar structures or function groups from the references. As demonstrated in the above experiments, our stochastic molecular generator proves to be effective in achieving the goal. Specifically, we aim at designing molecules that are effective on several critical opioid receptors including MOR, KOR, and DOR. The generated compounds are found to share similarities with the selected reference compounds. To generate novel molecules with desired binding affinity and drug-like properties, there are several concerns in utilizing our generator.

		\subsubsection{Binding affinity prediction reliability analysis}
				
		We incorporated two layers of BA predictors into our approach. The first layer, AE-BP, is utilized for initial BA screening, while the second layer consists of a consensus BA predictor, which provides more accurate predictions. This two-layered approach helps us identify promising drug candidates.
		
		However, considering the inherent nature of machine learning predictions, we anticipate that the generated potent compounds will exhibit certain similarities to the available training data. We carefully select reference compounds from the training data with BA values ranging from -10 kcal/mol to -12 kcal/mol. Within the three inhibitor datasets we collected, a considerable number of compounds fall within this specific BA range. If the generated compounds share certain degree of similarities with the training data, the machine learning models can effectively differentiate these potent compounds from the inactive ones.

		\subsubsection{Molecular novelties}
		
		The molecular novelties are assessed by measuring their similarity to selected reference compounds. The generator can effectively manage the level of novelty by adjusting the number of references or assigning weights to each reference compound. Higher molecular novelties indicate a broader coverage of the chemical space by the generated compounds, as they are different from the reference compounds or the available molecules in the training data. A wide range of chemical space, encompassing high binding affinity compounds, is advantageous for drug design. This enables a greater diversity in various druggable properties, such as Log P, Log S, Cano-2, and others. Having more drug candidates with different druggable properties provides a wider range of options for treating patients with varying health conditions. However, it is important to control the novelties of the generated compounds. Higher molecular novelties imply low similarities with the reference compounds or the machine learning training data. To ensure accurate binding affinity predictions, a certain degree of similarity with the available training data is still necessary.

		\subsubsection{Importance of selecting appropriate references}

		When generating novel molecules, we appropriately choose reference compounds from the training data that exhibit the desired potency on specific receptors. Our experiments have demonstrated that appropriate references are beneficial in generating a greater number of candidates with desired BAs. However, it is essential for a promising drug candidate to also meet other crucial drug-like properties, particularly the ADMET criteria.
				
		By utilizing reference compounds that demonstrate satisfactory ADMET properties, our generator can generate a substantial pool of compounds that exhibit the desired ADMET characteristics. This is exemplified in the analog generations for naltrexone and naloxone. Conversely, when the reference compounds do not meet the essential ADMET criteria, even millions of compounds are generated, they can hardly pass the ADMET screening process. This observation is illustrated in the demonstration example depicted in Figure \ref{Fig:example-optimal}. In this particular example, the majority of the generated compounds do not successfully pass the FDAMDD screening due to the utilization of reference compounds with a low FDAMDD profile.
		
		\subsubsection{Agonist/antagonist for opioid receptors}

		Approved medications, as mentioned earlier, function as agonists or antagonists on the opioid receptors, playing crucial roles in the treatment of OUD. Methadone and buprenorphine serve as examples of agonist medications that target MOR. Methadone, a long-acting MOR agonist, effectively alleviates withdrawal symptoms and cravings \cite{fareed2010effect}. On the other hand, buprenorphine acts as a partial agonist of MOR, producing milder effects compared to full agonists \cite{bickel1995buprenorphine}. Antagonist medications, in contrast, block opioid receptors, thereby preventing the binding of opioids and reducing their reinforcing effects. Buprenorphine, for instance, acts as a KOR antagonist, offering mood improvements for individuals with OUD \cite{mattick2014buprenorphine}. Naltrexone and naloxone are antagonists that target all three critical opioid receptors. In particular, naloxone exhibits the highest affinity for MOR and is used to counteract the respiratory and mental depression effects of opioid overdose \cite{algera2019opioid}.

		Recent advancements in deep generative research have introduced innovative approaches for the de novo design of improved opioid antagonists \cite{salas2023novo}, as well as the design for selective KOR antagonists \cite{deng2020towards}, recognizing the crucial role of antagonists in OUD treatment. Besides, machine learning models have been proposed to predict the agonist or antagonist activities of small molecules on MOR, KOR, and DOR \cite{sakamuru2021predictive}. These models offer the potential to prioritize compounds from extensive libraries for subsequent experimental testing.
		
		While it is essential to investigate the agonist/antagonist properties of our generated drug-like compound for OUD treatment, including initial machine learning predictions or experimental validations, this study does not encompass that scope. However, we plan to conduct such investigations in the future, either through collaborations with experimentalists or by constructing reliable machine learning predictive models

			\section{Conclusion}
						
			We have developed a highly effective deep generative model for generating novel molecules that can be effective on multiple targets, including MOR, KOR, and DOR. The molecular generator is designed by integrating a stochastic differential equation (SDE)-based diffusion approach into the latent space of a pretrained autoencoder model. Through careful selection of appropriate reference compounds and adherence to a series of novelty criteria, a substantial number of novel compounds with desirable binding affinities for MOR, KOR, and DOR, as well as other drug-like properties, can be generated.
			
			To predict the binding affinities, we employ advanced machine learning models that integrate autoencoder embeddings, transformer embeddings, and topological Laplacian fingerprints with machine learning algorithms. The incorporation of these diverse molecular representations enhances the accuracy of the binding affinity predictions. The selection of reference compounds is crucial in two aspects. Firstly, the number of reference compounds influences the novelty of the generated molecules. Secondly, the use of reference compounds with desired ADMET properties increases the likelihood of generating compounds that satisfy the necessary ADMET requirements.
			
			Extensive experiments have demonstrated the effectiveness of our deep generative models in designing molecules that exhibit structural similarities to known opioid molecules or alternative compounds with therapeutic potential. We utilized our generative network complex to generate a diverse set of drug-like molecules, but further experimental studies are needed to evaluate their pharmacological effective for OUD treatment. Our machine learning platform represents a valuable tool in addressing the urgent need for medications in the treatment of OUD. Additionally, our platform has the potential to facilitate the design of molecules that require specific selectivity on multiple targets, making it a promising tool for medication development in various diseases.

			\section*{Data availability}
			
			The related datasets studied in this work are available at: 
			https://weilab.math.msu.edu/DataLibrary/2D/. 
			
			\section*{Supporting Information}
			
			The Supporting information includes S1 Datasets and model performance summary, S2 Molecular similarity analysis, S3 ADMET indexes and the optimal ranges, S4 Additional drug-like compounds, S5 Additional analogs of approved medications, and S6 Element-specific topological Laplacian.

			\section*{Acknowledgment}	
			This work was supported in part by NIH grants R01GM126189 and R01AI164266, NSF grants
			DMS-2052983, DMS-1761320, DMS-2245903,  and IIS-1900473, NASA grant 80NSSC21M0023, MSU Foundation, Bristol-Myers Squibb 65109, and Pfizer.

			%
			%\renewcommand\refname{}
			%
			%% Use alpha to check for repeated references
			%\section*{References}
			%\bibliographystyle{unsrt}
			%\bibliography{gnc-opioids}

			% \end{multicols}
	\end{document}